\documentclass{article} %
\usepackage[preprint]{icml2020}  %
\usepackage{icml2020}  %

\usepackage{amsmath,amsfonts,bm}

\def\eqref#1{equation~\ref{#1}}

\def\1{\bm{1}}

\DeclareMathAlphabet{\mathsfit}{\encodingdefault}{\sfdefault}{m}{sl}
\SetMathAlphabet{\mathsfit}{bold}{\encodingdefault}{\sfdefault}{bx}{n}

\usepackage[utf8]{inputenc} %
\usepackage[T1]{fontenc}    %
\usepackage{hyperref}       %
\usepackage{url}            %
\usepackage{booktabs}       %
\usepackage{amsfonts}       %
\usepackage{nicefrac}       %
\usepackage{microtype}      %
\usepackage{appendix}
\usepackage{amsmath}
\usepackage{amssymb}
\usepackage{ulem}  %

\usepackage{todonotes}
\usepackage{subcaption}
\usepackage{bm} %
\usepackage{enumitem} %

\usepackage{mlmacros} %
\usepackage[nameinlink,capitalise]{cleveref} %

\presetkeys{todonotes}{%
  backgroundcolor=blue!10!white,
  linecolor=blue!10!white,
  bordercolor=blue!10!white
}{}

\usepackage[skip=5pt,font=small]{caption}
\usepackage{subcaption}

\usepackage{titlesec} %

\variables{l,e,s,x,y,z,b,k,g,y,n,w,D,P,S}
\variables[mean]{\mu}
\variables[std]{\sigma}

\probdists{p,q}

\definecolor{pink}{HTML}{ff00ff}
\definecolor{orange}{HTML}{ff9900}
\definecolor{blue}{HTML}{0000ff}

\newcommand{\render}{{\footnotesize \operatorname{\texttt{render}}}}

\renewcommand{\emph}{\textit}
\usepackage[acronym,smallcaps,nowarn,section,nogroupskip,nonumberlist]{glossaries}

\glsdisablehyper %

\newacronym{GAN}{gan}{generative adversarial network}
\newacronym{VAE}{vae}{variational auto-encoder}
\newacronym{IS}{is}{importance sampling}
\newacronym{IWAE}{iwae}{importance-weighted auto-encoder}

\newacronym{ELBO}{elbo}{evidence lower bound}
\newacronym{KL}{kl}{Kullback-Leibler}

\newacronym{MLP}{mlp}{multilayer perceptron}
\newacronym[firstplural=convolutional neural networks, plural=CNNs]{CNN}{cnn}{convolutional neural network}
\newacronym[firstplural=recurrent neural networks, plural=RNNs]{RNN}{rnn}{recurrent neural network}
\newacronym{GRU}{gru}{gated recurrent unit}
\newacronym{LSTM}{lstm}{long short-term memory}

\newacronym{REINFORCE}{reinforce}{REINFORCE}

\newacronym{SGA}{sga}{stochastic gradient ascent}
\newacronym{SGD}{sgd}{stochastic gradient descent}

\newacronym{ADAM}{adam}{ADAM}
\glsunset{ADAM}
\newacronym{RMSprop}{rmsprop}{RMSprop}
\glsunset{RMSprop}

\glsunset{REINFORCE}
\newacronym{RL}{rl}{reinforcement learning}

\newacronym{ELU}{elu}{exponential linear unit}

\newacronym{MNIST}{mnist}{mnist}
\glsunset{MNIST}

\newacronym{NERF}{n\textup{e}rf}{Neural Radiance Fields}
\newacronym{GQN}{gqn}{Generative Query Network}
\newacronym{GQN_baseline}{conv-ar-vae}{convolutional autoregressive \textsc{vae}}
\newacronym{SRN}{srn}{Scene Representation Network}
\newacronym{NVAE}{nvae}{Nouveau \textsc{vae}}
\newacronym{OURS}{n\textup{e}rf-vae}{Amortized Generative \textsc{n\textup{e}rf}}

\newacronym{GECO}{geco}{Generalized \textsc{elbo} Constrained Optimization}
\newacronym{AIN}{ain}{Adaptive Instance Norm}
\newacronym{MSE}{mse}{mean squared error}
\glsunset{MSE}
\newacronym{NP}{np}{neural process}

\icmltitlerunning{NeRF-VAE: A Geometry Aware 3D Scene Generative Model}

\begin{document}
\twocolumn[
\icmltitle{NeRF-VAE: A Geometry Aware 3D Scene Generative Model}

\icmlsetsymbol{equal}{*}

\begin{icmlauthorlist}
\icmlauthor{Adam R.\ Kosiorek}{equal,dm}
\icmlauthor{Heiko Strathmann}{equal,dm}
\icmlauthor{Daniel Zoran}{dm}
\icmlauthor{Pol Moreno}{dm}
\icmlauthor{Rosalia Schneider}{dm}
\icmlauthor{So\v{n}a Mokr\'a}{dm}
\icmlauthor{Danilo J.\ Rezende}{dm}
\end{icmlauthorlist}

\icmlaffiliation{dm}{DeepMind, London}
\icmlcorrespondingauthor{ARK}{\texttt{adamrk@google.com}}
\icmlcorrespondingauthor{HS}{\texttt{strathmann@google.com}}

\icmlkeywords{Machine Learning, ICML}

\vskip 0.3in
]

\printAffiliationsAndNotice{\icmlEqualContribution} %

\begin{abstract}

    We propose \glsunset{OURS}\gls{OURS}, a 3D scene generative model that incorporates geometric structure via \gls{NERF} and differentiable volume rendering.
    In contrast to \gls{NERF}, our model takes into account shared structure across scenes, and is able to infer the structure of a novel scene---without the need to re-train---using amortized inference.
    \Gls{OURS}'s explicit 3D rendering process further contrasts previous generative models with convolution-based rendering which lacks geometric structure.
    Our model is a \textsc{vae} that learns a distribution over radiance fields by conditioning them on a latent scene representation.
    We show that, once trained, \gls{OURS} is able to infer and render geometrically-consistent scenes from previously unseen 3D environments using very few input images.
    We further demonstrate that \gls{OURS} generalizes well to out-of-distribution cameras, while convolutional models do not.
    Finally, we introduce and study an attention-based conditioning mechanism of \gls{OURS}'s decoder, which improves model performance.

\end{abstract}
\vspace{-1em}
\begin{figure*}
    \centering
    \begin{minipage}{\linewidth}
        \centering
        \includegraphics[width=\linewidth]{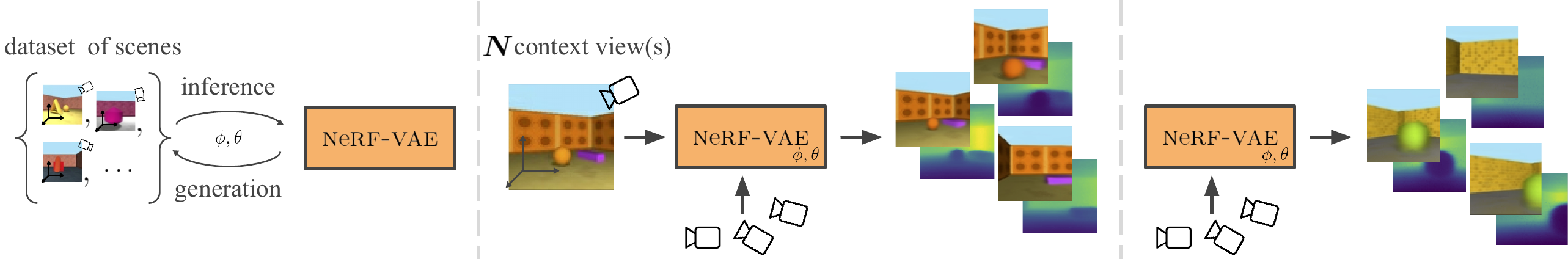}
    \end{minipage}
    \begin{minipage}{.29\linewidth}
        \centering
        \vspace{-1em}
        \subcaption{
        During \textbf{training}, we embed common scene properties (across the dataset) in the parameters $\phi, \theta$ of \acrshort{OURS}.
        }
        \label{fig:nerf_vae_fig_one_training}
    \end{minipage}
    \hfill
    \begin{minipage}{.40\linewidth}
        \centering
        \subcaption{
        Once the model is trained, we can \textbf{infer} parameters of a scene in constant time, even from a single view.
        }
        \label{fig:nerf_vae_fig_one_inference}
    \end{minipage}
    \hfill
    \begin{minipage}{.27\linewidth}
        \centering
        \subcaption{
        We can \textbf{sample} novel scenes from  \acrshort{OURS} generative model and render them from various viewpoints.
        }
        \label{fig:nerf_vae_fig_one_generation}
    \end{minipage}
    \vspace{-.75em}
    \caption{
        An overview of \gls{OURS}: a geometry-aware 3D scene generative model.
        \Gls{OURS} is trained on a dataset of several views (images and camera positions/orientations) from \textbf{multiple scenes}.
        Once trained, it allows for efficient inference of scene parameters (colours and geometry, including depth-maps) and sampling novel scenes from the prior.
    }
    \label{fig:nerf_vae_fig_one}
\end{figure*}

\section{Introduction}
\label{sec:intro}

The ability to infer the structure of scenes from visual inputs, and to render high quality images from different viewpoints, has vast implications for computer graphics and virtual reality.

This problem has traditionally been tackled with 3D reconstruction-based on matching visual keypoints, \eg \citet{lowe2004sift,schoenberger2016sfm}.
While these methods incorporate structure through multi-view geometry \cite{hartley2003mvg}, very few existing  methods adopt learned scene-priors such as types of objects or statistics of the background.
The resulting representations are usually discrete: they consist of meshes \cite{dai2017bundlefusion}, point clouds \cite{engel2014lsdslam}, or discretized volumes \cite{ulusoy2016patches}, and are difficult to integrate with neural networks.
An alternative approach to scene generation uses light-field rendering \citep{levoy1996light, SrinivasanWSRN17} and multiplane image representations \citep{ZhouTFFS18, MildenhallSCKRN19}.
While these methods allow image-based rendering, they do not estimate scene geometry. 
Hence, they are not directly useful for the purpose of 3D scene understanding.
More recent deep-learning-based novel view synthesis methods have the advantages of end-to-end training, and producing distributed representations that can be easily used in other neural-net-based downstream tasks.
These models, however, often have little embedded geometrical knowledge \cite{DosovitskiySTB17} and are either geometrically inconsistent \cite{nguyen2020bgan} or provide insufficient visual quality \cite{tatarchenko2015single}.
Moreover, many of these methods are deterministic and cannot manage uncertainty in the inputs \cite{sitzmann2019srn,trevithick2021grf}.

This work attempts to address these shortcomings. 
We introduce \gls{OURS}---a deep generative model with the knowledge of 3D geometry as well as complex scene priors.
Our work builds on \acrlong{NERF} (\acrshort{NERF}\glsunset{NERF}, \citet{mildenhall2020nerf}).
\Gls{NERF} combines implicit neural network representations of radiance fields, or \textit{scene functions}, with a volumetric rendering process.
\gls{NERF} needs to undergo a lengthy optimization process on many views of each single scene separately and does not generalize to novel scenes.
In contrast, \gls{OURS}---a \gls{VAE}---models multiple scenes, while its constant-time amortized inference allows reasoning about novel scenes.
Unlike \gls{NERF}, our model is also generative, and therefore capable of handling missing data and imagining completely new scenes.
In that, our work closely follows \acrlong{GQN}s (\acrshort{GQN}\glsunset{GQN}, \citet{eslami2018gqn}).
Like \gls{GQN}, \gls{OURS} defines a distribution over scene functions. 
Once sampled, a scene function allows rendering arbitrary views of the underlying scene.
Where \gls{GQN} relies on \glspl{CNN} with no knowledge of 3D geometry, leading to geometrical inconsistencies, \gls{OURS} achieves consistency by leveraging \gls{NERF}'s implicit representations and volumetric rendering.

\Gls{NERF} represents a scene by the values of \gls{MLP} parameters.
Being extremely high-dimensional, this representation precludes using amortized inference.
We change \gls{NERF}'s formulation to a scene function shared between scenes, and conditioned on a \textbf{per-scene} latent variable.
Intuitively, the latent variable captures scene-specific information (\eg position and kind of objects, colours, lighting, etc.), while shared information (\eg available textures and shapes, properties of common elements, sky)  is stored in the parameters of the scene function.
A prior over latent variables (and therefore scene functions) allows sampling novel scenes from the model, and rendering arbitrary viewpoints within them.
\Gls{OURS} parameters are learned using a collection of images from different scenes, with known camera position and orientation.
Since these parameters are shared between scenes (unlike in \gls{NERF}), \gls{OURS} can infer scene structure from very few views of an unseen scene.
This is in contrast to \gls{NERF}, where using few views results in poor performance. 
A high level overview of \gls{OURS} is depicted in \cref{fig:nerf_vae_fig_one}.

In summary, \gls{OURS} introduces four key benefits compared to existing models.
First, due to its amortized inference, it removes the need for costly optimization from scratch for every new scene.
Second, as it learns shared information between multiple scenes, it is able reconstruct unobserved scenes from a much smaller number of input views.
Third, compared to existing convolutional generative models for view synthesis, such as \gls{GQN},
it generalizes much better when evaluated on out-of-distribution camera views.
Finally, \gls{OURS} is the only amortized \gls{NERF} variant that uses compact scene representation in the form of a latent variable and that can handle uncertainty in the inputs.

\begin{figure*}
    \centering
    \includegraphics[width=\linewidth]{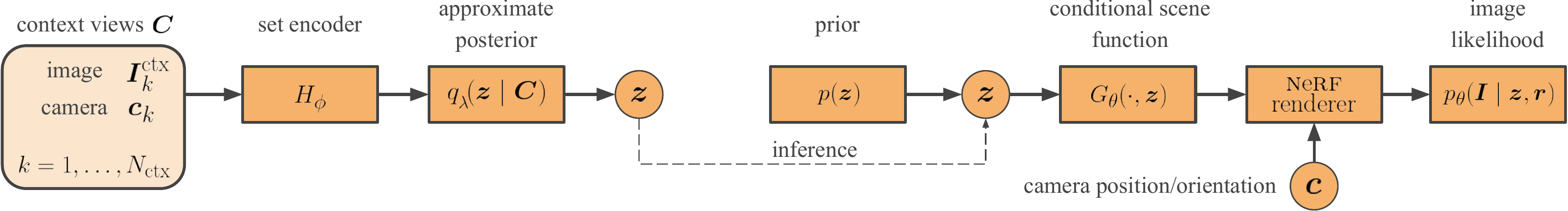}
    \begin{minipage}{.45\linewidth}
        \centering
        \subcaption{Inference in \gls{OURS}.}
        \label{fig:nerf_vae_inference}
    \end{minipage}
    \hfill
    \begin{minipage}{.44\linewidth}
        \centering
        \subcaption{The generative model of \gls{OURS}.}
        \label{fig:nerf_vae_gen_model}
    \end{minipage}
    \vspace{-.5em}
    \caption{Inference and generative model of \gls{OURS}. 
    For inference, a set $\bm{C}$ of context images $\bm{I}_k^\text{ctx}$ and cameras $\bm{c}_k$ from a scene are encoded into an approximate posterior distribution over the latent variable $\bm{z}$.
    This conditions a scene function $G_\theta(\cdot, \bm{z})$, which is used by the \gls{NERF} renderer to reconstruct images from arbitrary cameras within the scene.
    We can sample novel scenes by sampling the latent $\bm{z}$ from the prior.
    During training, the reconstruction \acrshort{MSE} and the \acrshort{KL} divergence are used in a variational optimization objective to learn the parameters $\theta$ of the conditional scene function and the parameters $\phi$ of the encoder.
    }
    \label{fig:nerf_vae_full}
\end{figure*}

\section{Neural Radiance Fields (\textsc{n}e\textsc{rf})}
\label{sec:nerf}
\Gls{NERF}'s scene function is represented as a 6D continuous vector-valued function whose inputs are ray coordinates $(\bm{x}, \bm{d})$ partitioned into position $\bm{x}\in\mathbb{R}^3$ and orientation\footnote{It is also possible to parameterize orientation using two angles, as \eg done by \citet{mildenhall2020nerf}, but we follow their implementation in using a vector.} $\bm{d}\in\mathbb{R}^3$.
Its outputs are an emitted colour $(r,g,b)\in\mathbb{R}^3$ and a volume density $\sigma \geq 0$.
The scene function is approximated by a neural network $F_\theta:(\bm{x}, \bm{d}) \mapsto ((r,g,b),\sigma)$ with weights $\theta$.
In order to encourage multi-view consistency, the architecture of this network is such that volume density $\sigma$ only depends on position while the emitted colour $(r,g,b)$ depends on both position and ray orientation.

Volumetric image rendering works by casting rays from the camera's image plane into the scene, one ray per pixel.
The colour that each ray produces is the weighted average of colours along the ray, with weights given by their accumulated volume densities.
\Gls{NERF}'s renderer uses a differentiable approximation to this accumulation process.
For details, we refer to \citet{mildenhall2020nerf, curless1996volumetric}.

We denote the image rendering process whose inputs are a \textbf{camera}  $\bm{c}$ (position and orientation) and a scene function $F_\theta$, and which outputs the rendered image\footnote{Note that it is easily possible to compute depth estimates for each rendered ray. We use this for visualization and evaluation purposes.} $\hat{\bm{I}}$, by 
\begin{align}
    \label{eq:vol_render}
    \hat{\bm{I}} = \render(F_\theta(\cdot), \bm{c})\,.
\end{align}
Note that computation of the camera's image plane and corresponding rays for each pixel requires camera parameters (\eg field of view, focal length, etc).

\section{Ne\textsc{rf-vae}}
\label{sec:gen_model}

    We build a generative model over scenes by introducing a view-independent latent variable $\bm{z}$ with prior $p(\bm{z})$ and a \textbf{conditional scene function} $G_\theta(\cdot, \bm{z}):(\bm{x}, \bm{d}) \mapsto ((r,g,b),\sigma)$.
    $G_\theta$ is much like \gls{NERF}'s $F_\theta$, but is additionally conditioned on $\bm{z}$, as will be detailed in \cref{sec:conditioning}.

    It is now the latent variable $\bm{z}$ that defines a \textbf{specific} scene, out of all the scenes that $G_\theta$ is able to represent, where $\theta$ are model parameters that capture  \textbf{shared structure} across scenes.
    The generative process of \gls{OURS} involves sampling $\bm{z}\sim\p(\bm{z})$ and using the resulting conditional scene function $G_\theta(\cdot, \bm{z})$ to render an image $\hat{\bm{I}} = \render(G_{\theta}(\cdot, \bm{z}), \bm{c})$ from a camera $\bm{c}$  with volumetric rendering from \cref{eq:vol_render}.
    
    The individual pixel colours $\hat{\bm{I}}(i,j)$ are used to define the mean parameter of a Gaussian image likelihood
    \begin{align*}
        p_\theta(\bm{I}\mid \bm{z}, \bm{c}) = \prod_{i,j}\mathcal{N}\left(\bm{I}(i,j) \mid \hat{\bm{I}}(i,j), \sigma^2_\text{lik}\right)
    \end{align*}
    with fixed or learned variance $\sigma^2_\text{lik}$, and we assume conditional independence of individual pixels $\bm{I}(i,j)$ given $\bm{z}$.

    Since the posterior over $\bm{z}$ is intractable, we approximate it---we frame \gls{OURS} as a \acrlong{VAE} (\acrshort{VAE}\glsunset{VAE}, \citet{kingma2013auto, rezende2014stochastic}) with \textbf{conditional} \gls{NERF} as its decoder.
    
\subsection{Amortized Inference for \gls{OURS}}
\label{sec:inference}
    Estimating the scene function parameters in \gls{NERF} is done separately for every scene, with no information sharing between scenes, which is  time-consuming, compute-intensive, and data-hungry.

    In contrast, \gls{OURS} introduces an encoder network $E_\phi$ with parameters $\phi$, which amortizes inference of the latent variable $\bm{z}$.
    Input for the encoder is a collection of $N_\text{ctx}$ \textbf{context views}, where each view consists of an image $\bm{I}^\text{ctx}\in\RR^{H\times W\times 3}$ and corresponding camera position and orientation $\bm{c}$.
    These views correspond to different viewpoints of a particular scene, forming a context set $\bm{C}:=\{\bm{I}_k^\text{ctx} , \bm{c}_k \}_{k=1}^{N_\text{ctx}}$.
    Each context element (concatenated camera $\bm{c}_k$ with image $\bm{I}_k^\text{ctx}$) is separately encoded using a shared encoder---we use a ResNet adapted from \citet{vahdat2020nvae}, details in \cref{app:arch_details}.
    The resulting $N_\text{ctx}$ outputs are subsequently averaged\footnote{
        We choose this method for its conceptual simplicity and note that more sophisticated options, such as attention-based mechanisms, might lead to better results as in \citet{trevithick2021grf}---exploration of which we leave for future work.
    } and mapped to parameters $\lambda$ of the approximate posterior distribution $\operatorname{q_{\hspace{0em}_\lambda}}\!(\bm{z}\mid\bm{C})$ over the latent variable $\bm{z}$.
    We use diagonal Gaussian posteriors.
    
    We fit the parameters $\{\theta, \phi\}$ of the \gls{OURS} by maximizing the following \gls{ELBO} on images,
    \begin{equation}
        \label{eq:elbo}
        \begin{aligned}
        &\loss[\gls{OURS}](\bm{I}, \bm{c}, \bm{C}; \theta, \phi) =\\
        & \mathbb{E}_{\bm{z} \sim q} \left[\log p_\theta \left(\bm{I}\mid\bm{z}, \bm{c} \right)\right]
        - \kl{\operatorname{q_{\hspace{0em}_\lambda}}\!\!\left(\bm{z}\mid\bm{C}\right)}{p(\bm{z})}\,.
        \end{aligned}
    \end{equation}
    In practice, we approximate the \gls{ELBO} by uniformly subsampling pixels from each image, see \cref{app:elbo_discuss} for details.

    \paragraph{Iterative Amortized Inference}
    Amortized inference suffers from \textit{amortization gap} \cite{cremer2018agap}---contrasting the gradient-based learning in \gls{NERF}.
    To bridge this gap, we employ iterative (amortized) inference \citep{kim2018sava, marino2018iterative}, which trades-off additional compute for improved inference.
    
    Iterative inference starts with an arbitrary guess for the posterior parameters, \eg $\lambda_0 = \bm{0}$, and iteratively refines them.
    At each step $t$, a latent is sampled from the current posterior $\bm{z}\sim q_{{\hspace{0em}_\lambda}_t}\!(\bm{z}\mid\bm{C})$.
    The sample is then used to render an image and to evaluate the gradient of the \gls{ELBO} in \cref{eq:elbo} \wrt $\lambda_t$.
    The gradient is passed to a recurrent refinement network (\textsc{lstm} followed by a linear layer)\footnote{
        Note that in this case, $\phi$ only contains parameters of the context encoder $E_\phi$ as well as the refinement network.
    } $f_\phi$, which updates the posterior parameters $\lambda_t$ for a given $\bm{C}$:
    \begin{align*}
        \bm{z}_t & \sim q_{{{\hspace{0em}_\lambda}_t}}\!\!\left(\bm{z} \mid \bm{C} \right)\,,\\
        {\lambda}_{t+1}&\leftarrow \lambda_t + f_\phi\left(E_\phi(\bm{C}), \nabla_{{\lambda}_t} \loss[\gls{OURS}]\right)\,.
    \end{align*}
    See \cref{app:arch_details} for further details.
    
    We emphasize that \gls{OURS}'s decoder uses explicit geometric structure, which  consequently is used by iterative inference.
    While this is only in an implicit manner (through \gls{ELBO} gradients), it differs from a geometry-agnostic feed-forward encoder typically used with \gls{VAE}s.

\subsection{Conditioning the Scene Function}
\label{sec:conditioning}
    
    \begin{figure}
        \begin{tikzpicture}
            \node[anchor=south west,inner sep=0] (image) at (0,0) {\includegraphics[width=\linewidth]{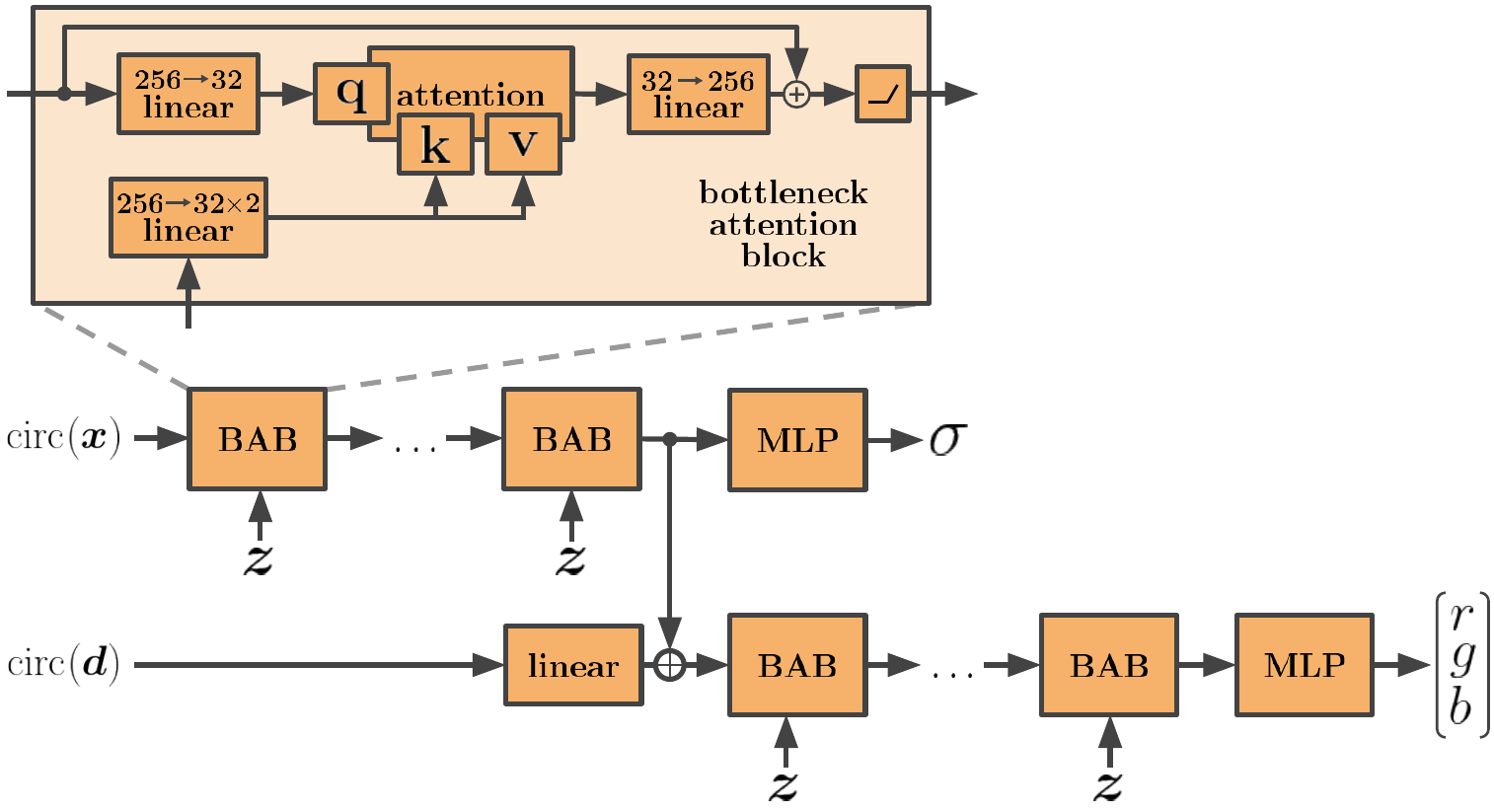}};
            \begin{scope}[x={(image.south east)},y={(image.north west)}]
                \node[anchor=north west] at (0.63,0.9) {\tiny 
                    $\operatorname{\text{circ}}(x) = \begin{pmatrix} \sin{(2^0 \pi x)}\\ \cos{(2^0 \pi x)}\\ \dots\\ \sin{(2^{L-1} \pi x)}\\ \cos{(2^{L-1} \pi x)} \end{pmatrix}$};
                \node[anchor=north west] at (0.68,0.55) {\tiny 
                    $\operatorname{\text{circ}}(\bm{x}) = \begin{pmatrix} \operatorname{\text{circ}}(x_1)\\ \dots\\ \operatorname{\text{circ}}(x_N) \end{pmatrix}$};
            \end{scope}
        \end{tikzpicture}
        \caption{
            Attention-based scene function. Input points $(\bm{x}, \bm{d})$, along rays corresponding to each pixel in the camera plane,  \textbf{attend} to different locations of the spatial latent variable $\bm{z}$ of size {\tiny $[H_\text{z} \times W_\text{z} \times D_\text{z}]$} using multi-head attention.
            The density part ($\sigma$)  depends only on the position $\bm{x}$ as in \gls{NERF}.
            Each attention block receives a different slice of the latent along its channel dimension.
            The inputs are projected to a lower-dimension space to save computation and memory;
            $\oplus$ is concatenation.
        }   
        \label{fig:att_scene_func}
    \end{figure}

    We now describe \gls{OURS}'s conditional scene function $G_\theta(\cdot,\bm{z})$, which is conditioned on the per-scene latent variable $\bm{z}$ and has additional across-scene parameters $\theta$.
    
    A simple way to condition \gls{NERF}'s scene function \acrshort{MLP} is to use $\bm{z}$ to shift and scale the inputs and activations at different layers---this mechanism resembles \textsc{ain} of \citet{dumoulin2017ain,brock2018large} and is referred to as an \gls{MLP} scene function henceforth.

    We further introduce an \textbf{attention-based} scene function, see \cref{fig:att_scene_func} for an overview.
    Attention \citep{vaswani2017attention} allows using a high-dimensional spatial {\footnotesize$[H_\text{z} \times W_\text{z} \times D_\text{z}]$ with} latent variable over which inputs of the scene function can attend.
    The spatial structure arises from removing the final average pooling across locations in the ResNet encoder.\footnote{Since the feature maps are averaged over context elements, locations in the latent do not necessarily correspond to parts of the scene; we use $H_\text{z} = W_\text{z} = 8$}
    Since the scene function is evaluated many times
    (\eg 256 in \citet{mildenhall2020nerf}; 128 in our experiments)
    for every pixel, it has to be computationally cheap with low memory footprint. 
    Consequently, we use only one linear layer per attention block and no layer norm.
    Additionally, our attention blocks are bottlenecks, i.e.\ keys, queries and values are low dimensional \cf
    \citet{srinivas2021bottlenecktransformer}.
    We find that these choices are a good trade-off between computation, memory and capacity, and used throughout our experiments.
    
    In order to model correlations between spatial locations, otherwise unaccounted for  by an independent prior $p(\bm{z})$, we apply a small \acrshort{CNN} to the latent before it is fed into the scene function, see \cref{app:arch_details} for further details.

\section{Related Work}
\label{sec:related_work}

\Gls{OURS} is closely related to amortized neural rendering approaches.
\citet{trevithick2021grf,yu2020pixelnerf} use \gls{NERF} as their decoder but require projecting all rendered points into the input space and thus do not formulate any compact scene representation.
\citet{tancik2020metanerf} suffers from a similar issue: it meta-learns good initializations for \gls{NERF}, but requires updating all parameters before it can render target observations.
\citet{sitzmann2019srn} uses sequential ray-marching instead of volumetric rendering, and demonstrates generalization only across objects from the same class.
\gls{OURS} has many similarities with \gls{GQN} of \citet{eslami2018gqn}, with the difference that \gls{GQN} uses a \gls{CNN} for rendering and is therefore not necessarily consistent across views.
\citet{mescheder2019occnet} show impressive single-image 3D reconstructions by modelling space occupancy with an implicit representation, though not modeling colours.

The above approaches work only when camera poses for input images are known.
This is not true for \textsc{gan}s, where it is enough to approximate the marginal distribution of poses in the training set if appropriate inductive biases are in place \citep{nguyen2019hologan}---an approach that also scales to an explicit multi-object setting \citep{nguyen2020bgan}.
\textsc{Graf} \citep{schwarz2020graf} and \textsc{giraffe} \citep{niemeyer2020giraffe} reuse the same idea for single- and multi-object settings, respectively, but use \gls{NERF} as the generator, which improves multi-view consistency.
Both approaches are related to \gls{OURS} in the sense that they use \gls{NERF} as a decoder which is conditioned on a latent variable---although \gls{OURS} introduces a more advanced conditioning mechanism.
Finally, in contrast to the \textsc{gan} approaches, \gls{OURS} has an associated inference procedure.

A limitation of \gls{NERF} is that it  can only model static scenes and does not support varying lighting conditions nor transient effects (\eg moving objects) often visible in the real world.
\Gls{NERF}-\textsc{w} \citep{martin2020nerf} addresses these shortcomings by adding per-view latent variables.
While our model makes latent variables explicit, it maintains a single latent per scene; an extension towards \textsc{nerf-w} is an interesting research direction.
\citet{pumarola2020dnerf,park2020nerfies,du2020nerf4d,li2020nerflow,xian2020spacetimenerf} further extend \gls{NERF} to dynamic scenes and videos: they formulate flow-fields that augment the original radiance field with a temporal component.
This complicated approach is necessary due to the high-dimensional nature of \gls{NERF}'s representation.
Since \gls{OURS} introduce a latent variable, it is possible to extend it to videos by simply adding a latent dynamics model (\eg \citet{hafner2020dreamer}).

\section{Experiments}
\label{sec:experiments}

        \begin{figure*}[h]
                \includegraphics[width=\linewidth]{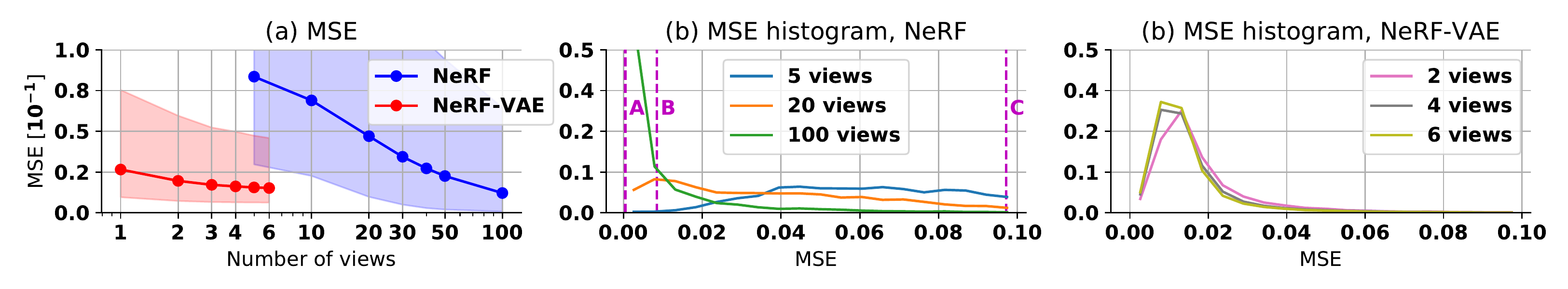}
                 \includegraphics[width=\linewidth]{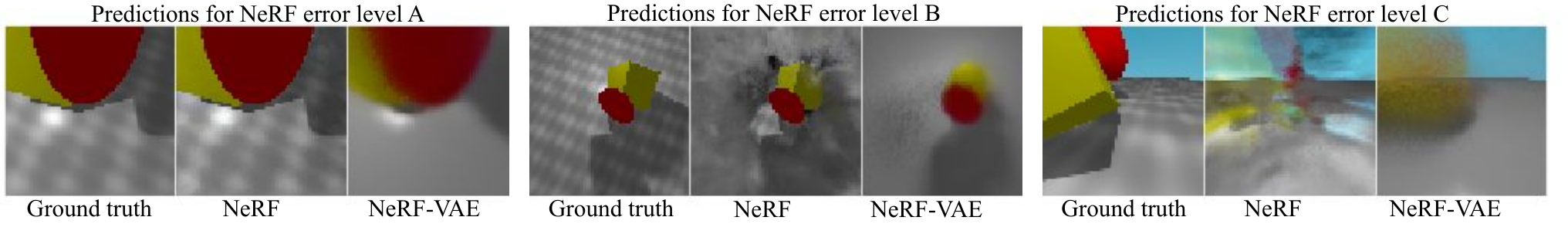}
                 \caption{
                Error analysis of \gls{NERF} and a minimalistic version (see text) of \gls{OURS} on Jaytracer data. \textbf{(a)}: \gls{MSE} decreases with increasing number of training (\gls{NERF}) and context (\gls{OURS}) views.
                We show the mean and 95\% percentiles across 100 test views averaged over 10 scenes and 10 seeds.
                Despite its minimal version, \Gls{OURS} performs much better for fewer ($\leq 6$) views; \gls{NERF} needs many more views ($\geq 100$) to reach comparable error.
                \textbf{(b, c)}:
                \Gls{MSE} histograms.
                Compared to \gls{OURS}, \gls{NERF} needs a large number of training views to consistently achieve low errors, and even then incurs a small number of larger errors.
                \textbf{Bottom}:
                An example scene, where views correspond to three error levels, indicated in (b), of \gls{NERF} trained on 100 views.
                \gls{NERF}'s predictions for level A are near perfect, but the model fails catastrophically for level C---which regularly happens when training on fewer views, see (b).
                \gls{OURS}'s predictions are more consistent, despite its simple decoder and inference.
            }
            \label{fig:exp1_errors_and_densities}
        \end{figure*}
        
        \begin{figure*}[h]
        \includegraphics[width=\linewidth]{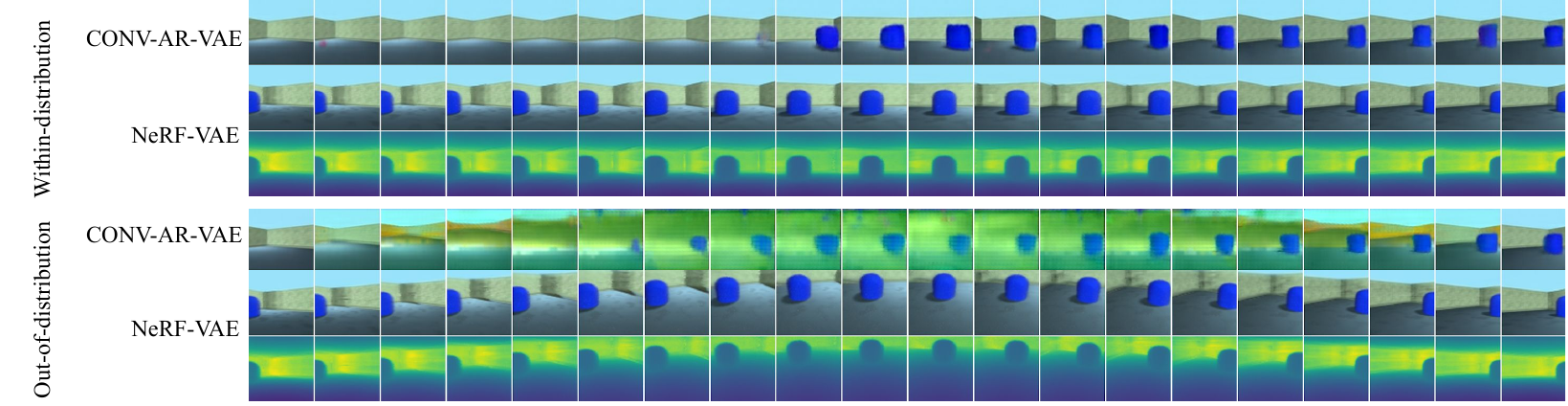}
        \caption{
            Generalization ability for interpolations along two trajectories.
            \textbf{Top}: the camera is moving parallel to the ground, facing the horizon (\textbf{within-distribution (WD)}, both models were trained using such views).
            While \gls{GQN_baseline} predicts the room consistently, it is inconsistent regarding the presence of the blue object.
            In contrast, \gls{OURS} produces fully consistent predictions, as further illustrated by visualizations of its inferred scene geometry.
            \textbf{Bottom}: the camera is further lifted off the ground, along a quadratic curve, facing downwards (\textbf{out-of-distribution (OOD)}).
            \gls{GQN_baseline} fails to account for downward camera orientations and outputs distorted colours, whereas \gls{OURS} produces plausible and consistent predictions and scene geometry.
        }
        \label{fig:exp2_out_of_plane_cameras}
        \end{figure*}

    To evaluate \gls{OURS}, we first
    analyze its ability to reconstruct novel viewpoints given a small number of input views, and contrast that with \gls{NERF}.
    Second, we compare our model with a \acrlong{GQN}-like autoregressive convolutional model, \citep[\gls{GQN}]{eslami2018gqn} %
    and show that while \gls{OURS} achieves comparably low reconstructions errors, it has a much improved generalization ability, in particular when being evaluated on camera views not seen during training.
    Third, we provide an ablation study of \gls{OURS} variants, with a focus on the conditioning mechanisms of the scene function described in \cref{sec:conditioning}.
    Finally, we showcase samples of \gls{OURS}.

    \subsection{Datasets}
        We use three datasets, each consisting of 64$\times$64 coloured images, along with camera position and orientation for each image, and camera parameters used to extract ray position and orientation for each pixel.
        \paragraph{GQN}
        \citep{eslami2018gqn}\footnote{We use the {\tiny\texttt{rooms\_free\_camera\_no\_object\_rotations}} variant publicly available at \url{https://github.com/deepmind/gqn-datasets}}, consists of 200k scenes  each with 10 images of rooms with a variable number of objects.
        Camera positions and orientations are randomly distributed along a plane within the rooms, always facing the horizon.
        Note that this dataset does not contain reflections or specularities, which we discuss in \cref{subsec:exp2}
        \paragraph{CLEVR}
        We created a custom CLEVR dataset \citep{Johnson_2017_CVPR} with 100k scenes, each with 10 views.
        Each scene consist of one to three randomly coloured and shaped objects on a plane.
        Camera positions are randomly sampled from a dome.
        Orientations are such that all objects are always present, making inference from a limited number of views easier than in the GQN data.
        The dataset is of higher visual quality than GQN, \eg it includes reflections.
        \paragraph{Jaytracer}
        In order to have more control over individual aspects of the data (\eg complexity, distribution of camera views, etc.),
        we also created a custom raytraced dataset which consists of 200k randomly generated scenes, each with 10 views, with a fixed number of objects in random orientations on a plane.
        Shapes, colours, textures, and light position are random, an example can be seen in \cref{fig:exp1_errors_and_densities}.
        
        For both CLEVR and Jaytracer, we generate 10 additional scenes each with 200 views and ground-truth depth maps for evaluation purposes and for training \gls{NERF}.
        See \cref{app:data_details} for more details.

    \subsection{Implementation details}

        The conditional scene functions' architecture follows \gls{NERF}, first processing position $\bm{x}$ to produce volume density and then additionally receiving orientation $\bm{d}$ to produce the output colours.
        Both position and orientation use circular encoding whereby we augment the network input values with a Fourier basis, \cf \cref{fig:att_scene_func}.
        
        We follow \citet[Section 5.2]{mildenhall2020nerf} in using hierarchical volume sampling in order to approximate the colour of each pixel.
        This means we maintain a second instance of the conditional scene function (conditioned on the same latent $\bm{z})$, which results in an additional likelihood term in the model log-likelihood in \cref{eq:elbo}.
        
        We use Adam \citep{kingma2014adam} and $\beta$-annealing of the KL term in \cref{eq:elbo}.
        Full details can be found in \cref{app:arch_details}.

    \subsection{Comparison with \textsc{N\textup{e}rf}}
        \label{subsec:nerf_comp}
        
        \gls{OURS}, unlike \gls{NERF}, can infer scene structure without re-training---through the  introduction of a per-scene latent variable and shared across-scene parameters.
        Our first experiment explores the following  questions:
        1.\ Can \gls{OURS} leverage parameter sharing to infer novel scenes from very few views?
        2.\ Is \gls{OURS}'s capacity affected by having a much smaller scene representation (a latent variable instead of an \gls{MLP}).
        3.\ At what number of views do both models reach comparable errors?

        To focus on these conceptual differences between \gls{NERF} and \gls{OURS}, we use \gls{OURS} with the simple \gls{MLP} scene function and without iterative inference. 
        We train \gls{OURS} on the Jaytracer data using $N_\text{ctx}=4$ context images (with corresponding cameras).
        We evaluate on 10 held-out scenes, with an increasing number of context views $N_\text{ctx}^\text{test}=1,\dots,6$.
        
        We train a separate instance of \gls{NERF} on each of these same 10 evaluation scenes, with an increasing number of training views $5, \dots, 100$.
        Both models are evaluated on images of 100 unseen views from the 10 evaluation scenes.
        We train and evaluate both models using 10 different seeds.
        See \cref{app:exp_details} for further training details.

    \subsubsection*{Results}
        \cref{fig:exp1_errors_and_densities} (a) shows reconstruction \glsunset{MSE}\gls{MSE}s (mean and 95\% percentiles) across the 100 test views.
        \Gls{OURS} achieves a significantly lower \gls{MSE} and uniformly lower worst case errors compared to \gls{NERF} trained on less than 100 views.
        This is despite \gls{OURS}'s amortized inference, which is many orders of magnitude faster than running the full optimization in \gls{NERF}.
        \Gls{NERF} achieves lower \gls{MSE} than our model only when sufficient training data is available (here $=100$ views).
        We emphasize that while \gls{OURS} was trained using $N_\text{ctx}=4$ context views, the model generalizes well to different numbers of context views at test time.

        Taking a look at the distribution of errors and the corresponding predictions in \cref{fig:exp1_errors_and_densities} (b, c), we see that \gls{NERF}'s \gls{MSE} distributions are extremely wide when trained on fewer than 100 views.
        Even for 100 views, \gls{NERF} suffers from a long tail of large errors.
        \Gls{OURS}'s errors concentrate on a small positive value for all evaluated $N_\text{ctx}^\text{test}$, with tails comparable to \gls{NERF} with 100 training views.

        \cref{fig:exp1_errors_and_densities}, bottom, shows predictions from both models on an example scene, for views corresponding to error levels attained by \gls{NERF} trained on 100 views: near-perfect (A), medium (B), and catastrophic (C), marked in  \cref{fig:exp1_errors_and_densities} (b).
        \gls{NERF}'s A-level predictions are near perfect, but significantly deteriorate in error levels B, C.
        As seen above, however, \gls{NERF} consistently (but not exclusively) attains level A only when trained with 100 views.
        We stress that in these simple scenes, this effect is not\footnote{
        Similar results, including catastrophic failures of \gls{NERF} trained with few views, can be obtained on CLEVR data where most of the scene is visible in all views, see \cref{app:exp1_details}.} due to incomplete scene coverage from a limited number of views---\gls{NERF} is simply not able to interpolate well between few training views.
        In contrast, \gls{OURS} captures the scene structure well on all error levels---though the simple version used here (\gls{MLP} scene function, no iterative inference) misses high frequency details such as sharp object boundaries and textures.
        This suggests that while amortized inference from few views is possible, \gls{OURS} could benefit from a more expressive scene function and a better inference mechanism.

    \subsection{Comparison with a Convolution-Based Generative Model}
    \label{subsec:exp2}
        Our next set of experiments compares \gls{OURS} to a \gls{GQN_baseline}---a model that is very closely related to \gls{GQN} of \citet{eslami2018gqn}, but with slight modifications to make it comparable to \gls{OURS}, see \cref{app:gqn_baseline} for details.
        
        Both \gls{OURS} and \gls{GQN_baseline} are able to infer novel scenes from few input images, and to subsequently synthesize novel views within those scenes.
        The key difference between the models is their decoder, responsible for rendering images given camera and latent scene representation: convolutional for \gls{GQN_baseline} and geometry informed for \gls{OURS}.

    \subsubsection*{Interpolations \& Generalization}
        We first illustrate that \gls{OURS}'s explicit knowledge of 3D geometry allows it to generalize well to out-of-distribution camera positions and orientations compared to \gls{GQN_baseline}.
        
        We train both \gls{OURS} and \gls{GQN_baseline} on the GQN dataset, using similar settings, which are detailed in \cref{app:exp2_details}.
        For the this experiment, we restrict \gls{OURS}'s scene function's access to orientations, which we will motivate below.
        For evaluation, we select a held-out scene and let each model infer the latent representation given the same $N_\text{ctx}^\text{test}=4$ views.
        We pick two arbitrary views and generate a sequence of camera positions and orientations that interpolates between them\footnote{Note that the GQN dataset does not contain ground truth images for these interpolated views.}.
        We do this twice: once with a simple linear interpolation across the plane, changing only the yaw of the camera (\textbf{within-distribution (WD) views}), and once with the camera lifting off the plane, along a quadratic curve, and looking down, changing both pitch and yaw (\textbf{out-of-distribution (OOD) views}); see \cref{fig:exp2_camera_trajectories} in \cref{app:exp2_details} for an illustration.
        We note that out-of-distribution here means that these camera positions lie outside the support of the training data distribution.
        
        \cref{fig:exp2_out_of_plane_cameras} shows each models' outputs along the within-distribution trajectories and out-of-distribution views trajectories.
        Both models produce plausible within-distribution interpolations, although \gls{GQN_baseline} has problems with object persistency.
        When evaluated on out-of-distribution views, \gls{GQN_baseline} completely fails to produce plausible outputs: ignoring the downward camera angle and instead distorting colours of the scene.
        In contrast, \gls{OURS} renders the inferred scene geometry properly from out-of-distribution viewpoints.
        \cref{fig:exp2_out_of_plane_cameras} further shows depth estimates for \gls{OURS}'s outputs, revealing the inferred scene geometry.
        We provide more examples and out-of-distribution experiments on CLEVR in \cref{app:exp2_details}.
        
    \begin{figure}
        \centering
        \centering
        \includegraphics[width=\linewidth]{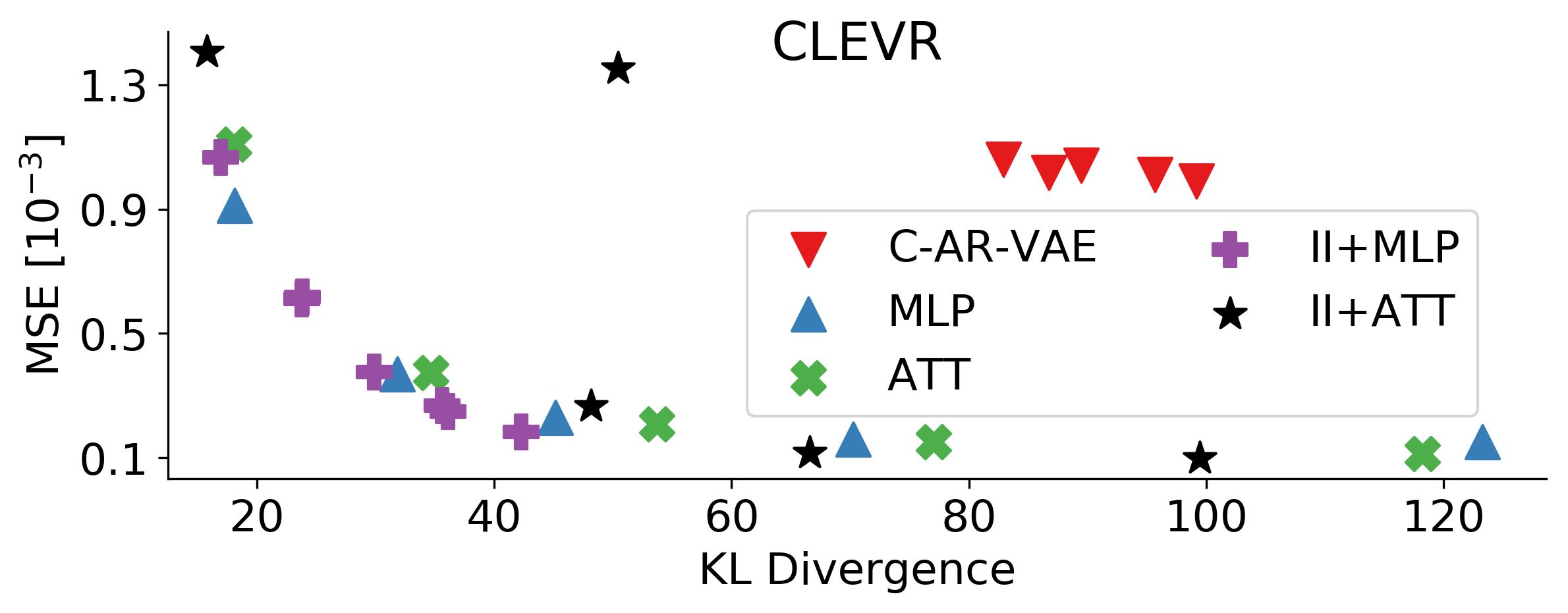}
        \includegraphics[width=\linewidth]{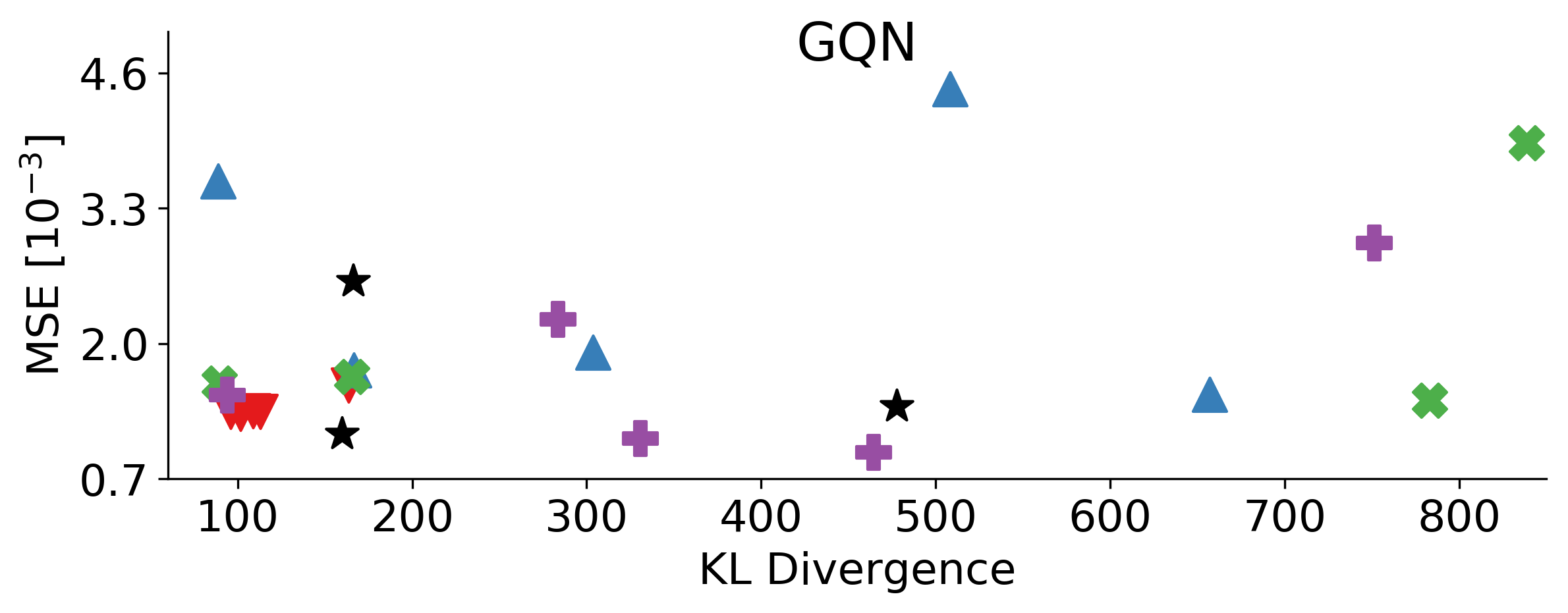}
        \caption{
            Variants of \gls{OURS} compared with \gls{GQN_baseline} on CLEVR and GQN datasets.
            We trained \gls{OURS} with a range of $\beta$ values to investigate reconstruction/\acrshort{KL} trade-offs.
            Iterative inference (\textsc{ii}) improves \gls{MLP}-based scene functions, but has a small effect on attentive (\textsc{att}) ones.
        }
        \label{fig:exp2_infoplane}
    \end{figure}
    
    \subsubsection*{Degenerate Orientations in the GQN data}
        Recall that camera views in the GQN data lie within a plane  parallel to the ground.
        Consequently, certain points in the scene are \textbf{only} observed from orientations within this plane.
        When trained on this degenerate data, evaluating a neural-network-based scene function on different (\eg when looking down) orientations inputs leads to unpredictable outputs.
        A simple way to circumvent the resulting rendering artifacts is to remove orientations from the scene function's inputs, as done in the above interpolations.
        In the case of the GQN data, which does not contain viewpoint dependent colours (reflections or specularities), this restriction does not impair visual output quality.
        We replicate the interpolations from \cref{fig:exp2_out_of_plane_cameras} with orientation inputs
        in \cref{app:gqn_cheating}.
    
    \subsubsection*{Quantitative Comparison \& Model Ablations}
    \label{subsec:ablations}
    
        \cref{fig:exp2_infoplane} shows \gls{MSE} and \gls{KL}-divergence of \gls{OURS} variants using amortized or iterative amortized (\textsc{ii}) inference and \gls{MLP} or attentive (\textsc{att}) scene functions, and juxtaposes them against the \gls{GQN_baseline}.
        We focus on CLEVR and GQN datasets.
        All \gls{OURS}s are trained with increasing values of $\beta$ to trace out the available trade-offs between \gls{MSE} and \gls{KL}.
        \Gls{GQN_baseline} is trained with \acrshort{GECO} \citep{rezende2018geco}, which
        allows to set a user-specified constraint on the likelihood term, details in \cref{app:exp2_details}.
        
        CLEVR contains simple objects visible from every viewpoint, and has no complicated textures.
        Consequently, the \gls{MLP} scene function achieves good \gls{MSE} and \gls{KL}.
        Iterative inference further increases reconstruction performance while decreasing \gls{KL}.
        It is interesting that \gls{MLP}-based models tend to have lower \gls{KL} values than attentive models while still maintaining good reconstruction.
        However, attentive models do obtain lower reconstruction errors in high-\gls{KL} regimes, which suggests that they have higher capacity to model complicated data.
        \gls{GQN_baseline} is not able to model the CLEVR data well---the model attains high \gls{KL} values despite manually setting a high \gls{MSE} threshold.

        The GQN dataset contains high-frequency textures and rooms which are not fully visible from every view, causing both a more difficult inference problem and more complicated rendering compared to CLEVR.
        The \gls{MLP} scene function without iterative inference achieves low \gls{KL} albeit generally higher \gls{MSE}.
        \textsc{att} can achieve lower errors than \gls{MLP} at similar \gls{KL} levels, despite also reaching very large \gls{KL} values, indicating a peaked posterior for certain values of $\beta$.
        Using iterative inference helps both decoders, and allows the model to achieve both lowest \gls{KL} and \gls{MSE}.
        We note that while \gls{GQN_baseline} achieves slightly lower \gls{MSE} and \gls{KL} on this within-distribution evaluation, it fails to generalize to out-of-distribution views as discussed above.

    \subsection{Samples \& Uncertainty}
        We now demonstrate \gls{OURS}'s capability to learn an unconditional prior over scenes (as opposed to images).
        
\begin{figure}[h]
        \includegraphics[width=\linewidth]{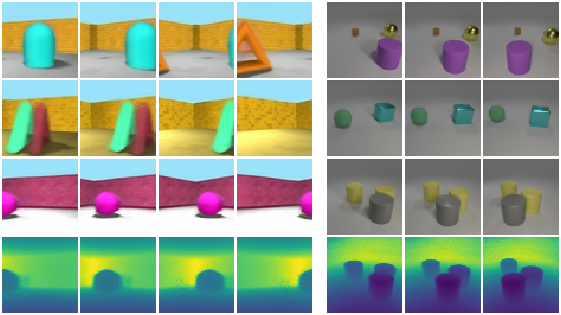}
        \caption{Sampled scenes from \gls{OURS}, trained on GQN (left) and CLEVR (right). Each row shows interpolated views in a different scene, corresponding to a sample of the latent variable. The last row shows a rendered depth map of the images in the above row (other depth maps omitted for space reasons). }
        \label{fig:exp_samples}
        \end{figure}

        We first sample the latent variable $\bm{z}\sim p(\bm{z})$, and then render a number of views of the induced scene function $G_\theta(\cdot, \bm{z})$.
        Samples from \gls{OURS} trained on the GQN and CLEVR datasets are shown in \cref{fig:exp_samples}.
        These samples resemble the training data distribution to a high degree, both in appearance and variability.
        Furthermore, the depth estimates (example in last row) reveal a consistent geometric structure of the sampled scene.
        
\begin{figure}[h]
        \includegraphics[width=\linewidth]{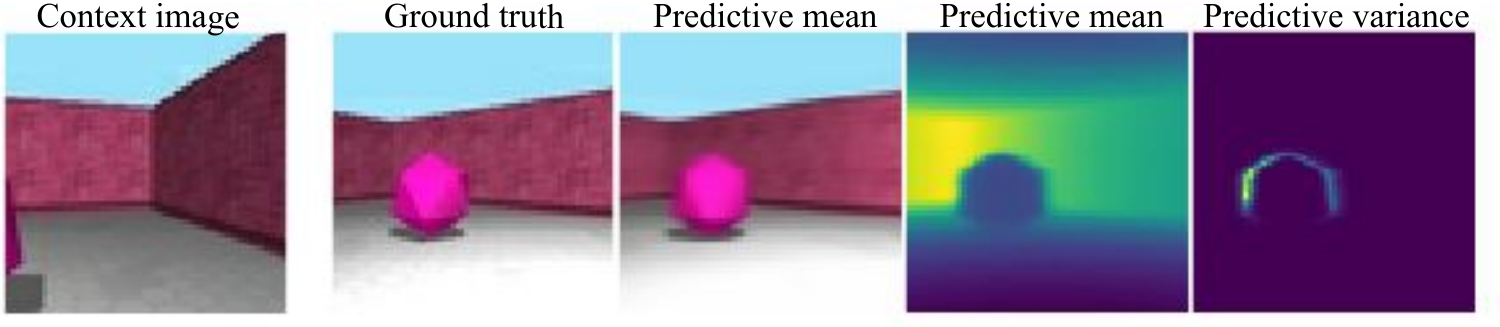}
        \caption{An example of predictions under limited information in the context view. We show colour and depth predictions. \gls{OURS} is able to reconstruct a plausible explanation of the object which is only marginally visible in the single context view. The predictive variance of depth estimates (averaged over 100 samples from the posterior over the latent variable) accounts for object boundaries not clearly visible in the context.}
        \label{fig:uncertainty}
        \end{figure}

        \cref{fig:uncertainty} shows an example where \gls{OURS} infers a scene from a single context image containing a barely-visible pink object.
        The model is able to accurately predict a view that contains the full object.
        At the same time, the model maintains uncertainty about the exact shape (sphere vs.\ icosahedron), as can be seen in the predictive variance of depth estimate.

\section{Discussion}
\label{sec:conclusion}

We presented \gls{OURS}, a geometry-aware scene generative model that leverages \gls{NERF} as a decoder in a \gls{VAE}-framework.
Thanks to an explicit rendering procedure, \gls{OURS} is view-consistent and generalizes to out-of-distribution cameras, unlike convolutional models such as \gls{GQN}.
Additionally, the learned prior over scene functions allows \gls{OURS} to infer scene structure from very few views.
This is in contrast to \gls{NERF}, where too few views may result in contrived explanations of a scene that work for some views but not for others.
The combination of geometric structure and a prior over scene functions, however, is not a panacea: if the data is degenerate (\eg cameras restricted to a plane) and the model is overparameterized (\eg unnecessarily accounting for view-dependent colours), \gls{OURS} might still explain the data in implausible ways.

One of the limitations of \gls{OURS} is its limited per-scene expressivity.
\gls{NERF} allocates \textbf{all} of its capacity to a single scene, and is able to capture high levels of complexity.
\Gls{OURS}, however, splits its capacity between shared across-scene information (conditional scene function) and per-scene information (latent).
In order to allow for amortized inference, the capacity of the latent needs to be limited---resulting in reduced per-scene expressivity compared to \gls{NERF}.

At the same time, a low-dimensional latent variable opens  interesting possibilities for  future work, \eg interpolation between scenes, extensions to dynamic scenes and videos, and a latent representation that dynamically grows with input complexity.

\newpage
\bibliography{library}
\bibliographystyle{icml2020}

\newpage
\onecolumn %
\appendix
\icmltitle{NeRF-VAE: A Geometry Aware 3D Scene Generative Model\\Supplementary Material}
\section{Dataset Details}
\label{app:data_details}

    \begin{center}
        \includegraphics[width=.49\linewidth]{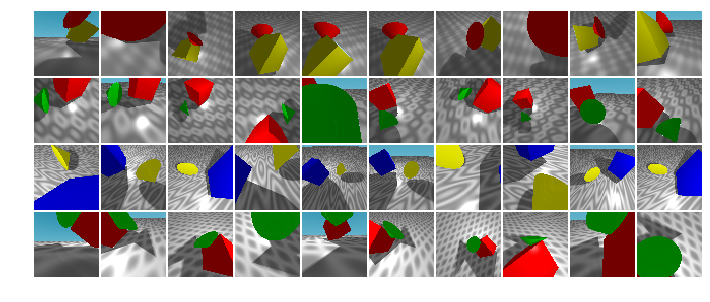}\hspace{0.1cm}
        \includegraphics[width=.49\linewidth]{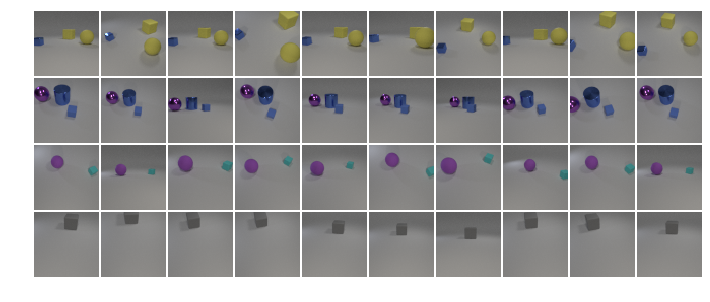}
        \captionof{figure}{
            Multiple views (columns) of example scenes (rows) in the Jaytracer and CLEVR datasets. \textbf{Jaytracer} (left): This dataset has a much more diverse distribution of camera perspectives than both GQN and CLEVR---making scene inference  more challenging. The example predictions in \cref{fig:exp1_errors_and_densities} are from the first scene shown here.
            \textbf{CLEVR} (right): Camera positions, angles and number of objects are randomized, compared to the original CLEVR data.
        }
        \label{fig:jaytracer_examples}
    \end{center}

    \subsection{CLEVR} CLEVR \citep{Johnson_2017_CVPR} is a widely used synthetic image dataset designed for visual reasoning. The generative process of a scene involves placing a number of objects of different shapes (cube, sphere, and cylinder), with randomly selected material properties, sizes, and positions on a ground plane, and random light positions.
    The camera is placed on a fixed location, $(7.48, -6.50, 5.34)$, with a small amount of random jitter added.
    We make a number of slight modifications that involve randomizing the camera pose and number of objects:
    \begin{itemize}
        \item The camera's distance to the origin is uniformly sampled between $[6, 8]$,
        \item the elevation angle is shifted by an angle uniformly sampled between $[-20, 45]$,
        \item the azimuth angle is shifted by an angle uniformly sampled between $[-15, 40]$ degrees.
        \item Each scene is instantiated with one, two or three objects.
    \end{itemize}
    Examples are shown in \cref{fig:jaytracer_examples}

    \subsection{Jaytracer}
    The Jaytracer dataset was rendered using a customized raytracer written in Jax\footnote{\url{http://github.com/google/jax}}.
    All scenes consist of an infinite ground plane, a dome acting as sky, and two objects.
    Ground textures are randomly generated for each scene, using a simple series linear operations and periodic nonlinearities, and a random colour within a range of gray-scales.
    The sky colour is random within a fixed range.
    Objects shapes and colours are randomly chosen from a fixed set (torus, box, sphere, cylinder, a set of 5 distinct colours), are randomly scaled within a given range, and randomly positioned within a rectangle around (-1,1) in the ground plane.
    Camera positions are uniformly sampled from a dome and oriented towards the origin.
    There is a single point source of white light, which is randomly placed within a box of a fixed size, along with a fixed level of ambient light.
    Scenes are represented as a collection of signed distance functions.
    We use a Lambert and a Blinn Phong Shader with a single pass of shadow rendering.
    Examples are shown in \cref{fig:jaytracer_examples}

\section{Architectural Details}
\label{app:arch_details}
    \subsection{Hierarchical sampling in \gls{NERF}}
    \label{app:subsec:nerf_details}

    \gls{NERF} uses hierarchical sampling in the rendering process.
    First, a `coarse' scene function is evaluated at various points along the rays corresponding to each image pixel.
    Second, based on the density values at these coarse points, a second set of points is re-sampled along the same rays, and a second `fine' scene function (with the same architecture) is evaluated at the re-sampled points.
    The resulting (fine) densities and colours are used in \gls{NERF}'s volume rendering mechanism.
    In order to enable gradient updates of the coarse scene function (re-sampling implementation not differentiable), \gls{NERF} reconstructs pixel colours using both scene functions' output, and minimizes a sum of the coarse and fine pixel errors.
    We follow this approach in all decoders of \gls{OURS}.
    This means that we maintain a coarse and a fine instance of the scene function network, and our likelihood in practice is
    \begin{align*}
        p_\theta(\bm{I}\mid \bm{z}, \bm{c}) =& \prod_{i,j}\mathcal{N}\left(\bm{I}(i,j) \mid \hat{\bm{I}}_\text{coarse}(i,j), \sigma^2_\text{lik}\right)\\
        \times& \prod_{i,j}\mathcal{N}\left(\bm{I}(i,j) \mid \hat{\bm{I}}_\text{fine}(i,j), \sigma^2_\text{lik}\right)
    \end{align*}
    where $\hat{\bm{I}}_\text{coarse}(i,j), \hat{\bm{I}}_\text{fine}(i,j)$ are the rendered images before and after the re-sampling step described above.

    \subsection{Encoder}
    \label{app:encoder_details}
    
    \begin{center}
        \begin{minipage}{0.49\linewidth}
            \includegraphics[width=\linewidth]{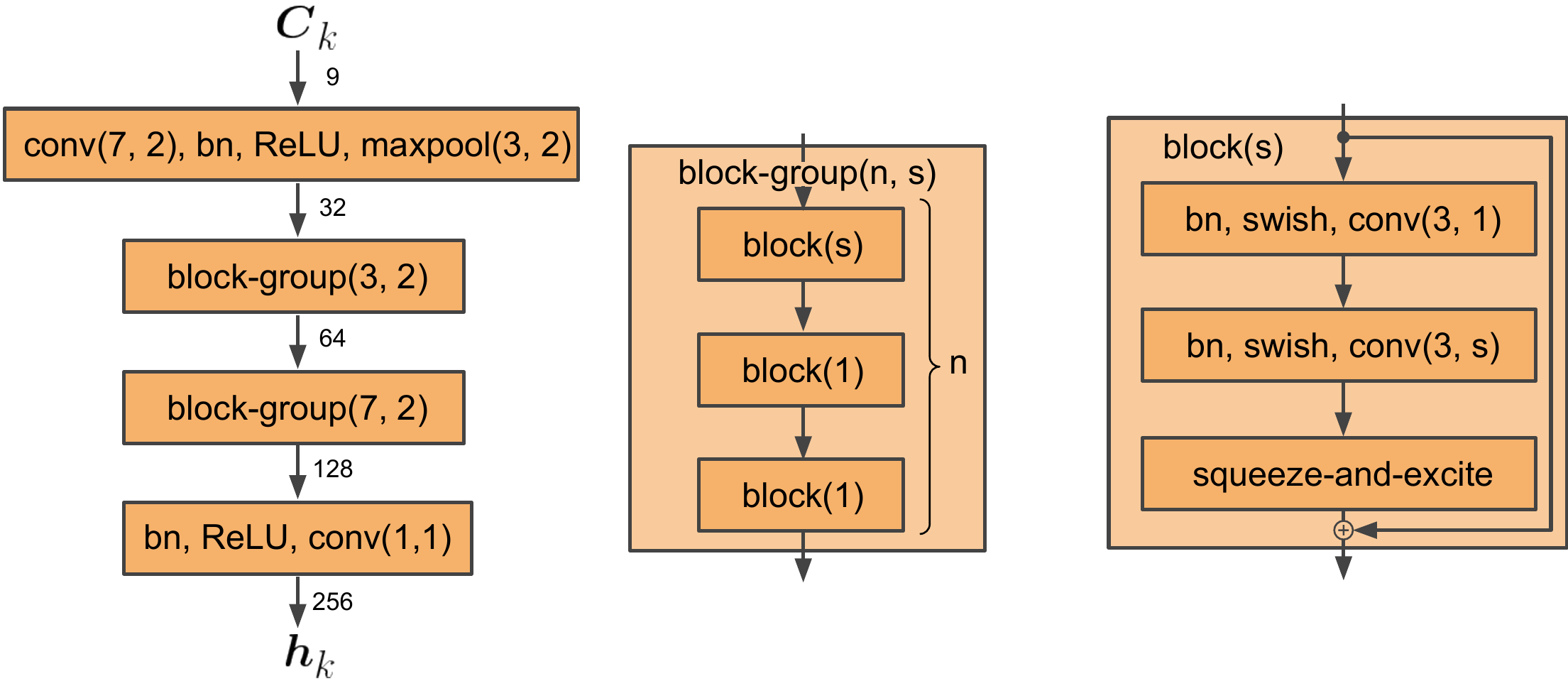}
        \end{minipage}
        \hfill
        \begin{minipage}{0.49\linewidth}
            \captionof{figure}{
                NouveauResNet adapted from \citet{vahdat2020nvae} and based on the ResNet v2 of \citet{he2016resnetv2};
                bn stands for batch norm; ReLu and swish are activation functions.
                The arguments are (kernel size, stride) in conv$(\cdot,\cdot)$ and maxpool$(\cdot,\cdot)$, the number of blocks and the stride of the first block in block-group$(\cdot,\cdot)$, and stride in block$(\cdot)$.
                The numbers of channels are given next to the arrows.
                Convolutions do not use additive biases.
            }
            \label{fig:nresnet}
        \end{minipage}
    \end{center}
    The approximate posterior $q_{\hspace{0em}_\lambda}\!(\bm{z}\mid\bm{C})$ is conditioned on the context $\bm{C} = \set{\bm{I}_k^\text{ctx},\bm{c}_k}_{k=1}^{N_\text{ctx}}$, which consists of images and associated camera position and orientation.
    We use the camera parameters to map $\bm{c}_k$ into a 6-dimensional feature map containing positions of each pixel in $\bm{I}$ (constant across one image), and the pixel's corresponding ray orientation.
    We express orientations as a 3-dimensional unit vectors.
    We then concatenate the image $\bm{I}_k^\text{ctx}$ with the positions and orientation feature map to form a context element $\bm{C}_k$.
    The context elements are then encoded with a NouveauResNet derived from \citet{vahdat2020nvae}, namely,
    \begin{equation}
    \begin{aligned}
        \bm{C}_k &= \texttt{concat}(\bm{I}_k^\text{ctx}, \texttt{map\_to\_rays}(\bm{c}_k))\,,\\
        \bm{h} &= \frac{1}{N_\text{ctx}}\sum_{k=1}^{N_\text{ctx}} \text{NouveauResNet}(\bm{C}_k)\,\\
        \lambda &= \textsc{mlp}(\bm{h})\,
    \end{aligned}
    \label{eq:encoder}
    \end{equation}
    where the final \gls{MLP} is applied element-wise if $\bm{h}$ is a feature map.
    Using $\lambda$, we obtain the posterior parameters by splitting $\lambda$ into two parts for mean and standard deviation, where we use a softplus to ensure positivity of the latter.
    
    The models that use the \gls{MLP} scene function require a vector-valued latent $\bm{z}$.
    In this case we apply global average pooling on $\bm{h}$ before it is fed into the \gls{MLP} in \cref{eq:encoder}. 
    Attentive models use a feature-map-sized latent and do not require pooling in the encoder.
    
    \paragraph{Iterative amortized inference}
    For iterative inference, we approximate the ELBO loss (\cref{eq:elbo}) by subsampling a batch of pixels from images in the context (see also discussion below), and using that to calculate the gradient of the loss w.r.t.\ the posterior parameters $\lambda$.
    This gradient, together with $\bm{h}$, are then fed input to an LSTM, whose outputs are linearly mapped and added onto the current estimate of $\lambda$.
    Posterior parameters are obtained (from the final estimate of $\lambda$) as above.

    \paragraph{Image encoder}
    Our implementation of NouveauResNet follows the ResNet v2 of \citet{he2016resnetv2} but uses swish activations and squeeze-and-excite (with the reduction ratio of 8, \citet{hu2020sen}) modules as done by \citet{vahdat2020nvae}.
    
    \subsection{Scene Functions}
    \label{app:scene_func_detail}
    
    \begin{table}
        \centering
        \begin{minipage}{0.49\linewidth}
        \centering
        {\small
        \begin{tabular}{c|c|c|c}
            \toprule
            Model & \textsc{elbo} & \textsc{mse} {\tiny $\left[10^{-3}\right]$} & \textsc{kl}\\
            \midrule
            Shift & $3.73$ {\tiny $\pm 0.051$} & $2.53$ {\tiny $\pm 0.344$} & $703$ {\tiny $\pm 12$}\\
            Shift All & $3.78$ {\tiny $\pm 0.007$} & $2.20$ {\tiny $\pm 0.047$} & $643$ {\tiny $\pm 8$}\\
            \textsc{Ain} All & $3.78$ {\tiny $\pm 0.010$} & $2.20$ {\tiny $\pm 0.063$} & $670$ {\tiny $\pm 6$}\\
            Att & $\mathbf{3.90}$ {\tiny $\mathbf{\pm 0.010}$} & $\mathbf{1.48}$ {\tiny $\mathbf{\pm 0.063}$} & $\mathbf{414}$ {\tiny $\mathbf{\pm 20}$}\\
            \bottomrule
        \end{tabular}}
        \subcaption{GQN data; scene function \textbf{without} orientation inputs.}
        \label{tab:exp2_ablations_without_orientation}
        \end{minipage}
        \begin{minipage}{.49\linewidth}
        \centering
        {\small
        \begin{tabular}{c|c|c|c}
            \toprule
            Model & \textsc{elbo} & \textsc{mse} {\tiny $\left[10^{-3}\right]$} & \textsc{kl}\\
            \midrule
            Shift & $3.84$ {\tiny $\pm 0.050$} & $1.76$ {\tiny $\pm 0.331$} & $692$ {\tiny $\pm 5$}\\
            Shift All & $3.87$ {\tiny $\pm 0.008$} & $1.59$ {\tiny $\pm 0.052$} & $661$ {\tiny $\pm 7$}\\
            \textsc{Ain} All & $3.88$ {\tiny $\pm 0.006$} & $1.54$ {\tiny $\pm 0.035$} & $660$ {\tiny $\pm 7$}\\
            Att & $\mathbf{3.97}$ {\tiny $\mathbf{\pm 0.027}$} & $\mathbf{1.06}$ {\tiny $\mathbf{\pm 0.172}$} & $\mathbf{345}$ {\tiny $\mathbf{\pm 16}$}\\
            \bottomrule
        \end{tabular}}
        \subcaption{GQN data; scene function \textbf{with} orientation inputs.}
        \label{tab:exp2_ablations_with_orientation}
        \end{minipage}
        \caption{
        Scene function ablations on the \gls{GQN} dataset.
        Using orientation inputs in the scene function significantly improves performance but leads to overfitting on this dataset, \cf \cref{app:gqn_cheating}.
        We try to equalize \gls{KL} of different models by setting different $\beta$ values in the \gls{ELBO} so that we can compare reconstruction errors directly.}
        \label{tab:exp2_scene_func_ablations}
    \end{table}

    \Gls{NERF} uses a \glsreset{MLP}\gls{MLP} as its scene function.
    Similarly, we can form a conditional scene function $G_\theta$ by conditioning an \gls{MLP} on $\bm{z}$ as follows.
    Concatenate $\bm{z}$ with the input point $(\bm{x},\bm{d})$ (Shift).
    This is equivalent to setting a scene-specific bias in the first layer of $F_\theta$.
    Concatenate $\bm{z}$ with the inputs to all layers of $G_\theta$ (Shfift All), which is equivalent to setting scene-specific bias vectors in all layers.
    Use $\bm{z}$ to shift and scale inputs to all layers of $G_\theta$ (\textbf{\textsc{ain} all}).
    This is similar to \gls{AIN} of \citet{dumoulin2017ain}, and translates to setting a scene-specific bias and performing column-wise scaling of the corresponding weight matrices of $G_\theta$.
    Or we can use attention (Att) as described in \cref{sec:conditioning}.
    We compare these four options on the GQN data using scene function that either \textbf{do} or \textbf{do not use orientation inputs}.
    The results are available in \cref{tab:exp2_scene_func_ablations}.
    We used different $\beta$ values in the \gls{ELBO} to achieve similar \gls{KL} values.
    Not surprisingly, increasing complexity of the conditioning mechanism leads to decreasing reconstruction errors, with the attentive scene functions of \cref{sec:conditioning} achieving the best results.
    
    \vspace{-1em}\paragraph{Latent variable pre-processing} We use a small \gls{MLP} or \gls{CNN} (in attentive scene functions) on the latent \textbf{before} it is fed into the scene function.
    Originally introduced to account for relations between locations in spatial latent variables as described in \cref{sec:conditioning}, this mechanism improves results in all models.
    
    \vspace{-1em}\paragraph{Fourier encoding} We use $L=10$ and $L=4$-component Fourier encoding for positions and orientations, respectively; see \cref{fig:att_scene_func} for details. 

\section{\Gls{GQN_baseline} Baseline Details}
\label{app:gqn_baseline}
    The \gls{GQN_baseline} baseline used in our experiments is closely related to \gls{GQN} of \citet{eslami2018gqn}.
    We begin by describing the variant of the graphical model used in our experiments and highlighting differences with the original model.
    \Gls{GQN_baseline} defines the following generative model
    \begin{equation}
        \begin{aligned}
        p_\theta(\bm{I}, \bm{z} \mid \bm{c}) = p_\theta(\bm{I} \mid \bm{z}) p_\theta(\bm{z} \mid \bm{c})
        \end{aligned}
    \end{equation}
    and the associated approximate posterior $q_\phi(\bm{z} \mid \bm{C}, \bm{c})$.
    The latent variable $\bm{z}$ is composed of multiple latents $\bm{z}^i, i=1,\dots,L$  generated in an autoregressive manner.
    Note that in contrast to our \gls{OURS}, both the prior and the posterior are conditioned on the camera $\bm{c}$, which results in a model that is \textbf{not} consistent across different views.
    This is because rendering an image from a different viewpoint requires re-sampling of the latent variable, and therefore changing the scene characteristics.
    The model \textbf{can} be consistent only when re-sampling from the posterior, when the posterior's uncertainty is minimized by providing sufficient information in the context $\bm{C}$.
    
    \vspace{-1em}\paragraph{Graphical model differences}
    In \gls{GQN_baseline}, the prior is conditioned on the camera but not on the context.
    The context is used only to condition the posterior.
    During training, we always reconstruct the same images that were provided in the posterior context $\bm{C}$, and we never condition the model on any additional images.
    The context \textbf{always} contains four distinct views of a scene.
    In contrast, \gls{GQN} used a prior and likelihood conditioned on the prior context $\bm{C}_\text{prior}$, while the posterior was conditioned on a context $\bm{C}_\text{posterior}$ that contained the $\bm{C}_\text{prior}$ as well as additional images, and was always trained to reconstruct a single image from the posterior context. 
    Both the prior and posterior contexts vary in size between minibatches, with occasionally empty prior context (in this case the model is equivalent to the baseline we used).
    The latter enables the model to sample scenes unconditionally.
    
    \vspace{-1em}\paragraph{Neural architecture of \gls{GQN_baseline}}
    The posterior $q_\phi$ uses the same per-image encoder $H_\phi$ as \gls{OURS}, see \cref{app:arch_details} for details, with the difference that it uses $[(5, 2), (5, 1)]$ block groups (see \cref{app:encoder_details}) for details.
    The prior and posterior both use an autoregressive process, implemented by a convolutional \textsc{lstm} with kernel of size $5$, $128$ hidden units and $16 \times 16$ spatial dimensions.
    At each stage, the \textsc{conv-lstm} takes as input the latent $\bm{z}_{i-1}$ from the previous stage concatenated with pre-processed camera (c.f.\ \cref{app:arch_details}) and the encoded context $\bm{C}$ for the posterior.
    For the prior, the distribution parameters are computed by a single convolution with kernel of size $3$ from the output of the prior \textsc{conv-lstm}.
    To compute the posterior, we take the prior distribution statistics, and we add a linear projection (from a \textsc{conv} of kernel size $3$) of the posterior \textsc{conv-lstm} outputs.
    All prior and posterior distributions are diagonal Gaussian.
    The latent variable at each stage is $16 \times 16 \times 16$, with $L=7$ stages.
    
    The latent variables are used to sequentially update a canvas of the same size as the image (in our case $64 \times 64 \times 3$).
    To do that, we use a transposed convolution with $\text{kernel size} = \text{stride} = 4$ and $128$ channels, followed by a convolution with kernel of size $3$ and $3$ output channels, whose output is added to the canvas.
    Finally, the canvas is mapped to the mean of a Gaussian distribution with a fixed scale parameter of $0.1$ by using a single per-pixel convolution.
    The model is trained with \gls{GECO} \cite{rezende2018geco}.

\section{Estimating the \Gls{ELBO}}
\label{app:elbo_discuss}
    \Gls{OURS}, despite presented as a model of images, assumes conditional independence of the individual image pixels given the latent variable.
    Consequently, the model can be viewed as estimating the conditional probability of a single pixel given its corresponding ray in the camera's image plane.
    This perspective is useful when discussing how to estimate the ELBO and to train with GECO, which we do below.
    
    \paragraph{From image to pixel likelihood}
    Let $\bm{c}(i)$ be a deterministic function of the camera $\bm{c}$ and the pixel index $i$, mapping pixel index $i$ to ray position and orientation (using the camera parameters, further discussed in \cref{app:encoder_details}).
    A single pixel joint likelihood then is $p_\theta(\bm{I}(i), \bm{z} \mid \bm{c}(i)) = p_\theta(\bm{I}(i) \mid \bm{z}, \bm{c}(i)) p(\bm{z})$.
    The resulting \gls{ELBO} is equivalent to $\beta$-\gls{ELBO} of image models when $\beta = \nicefrac{1}{\text{num pixels}}$.
    Therefore, we restore $\beta = 1$ by defining the likelihood term of \gls{OURS} as $p_\theta(\bm{I} \mid \bm{z}, \bm{c}) =  \sum_{i=1}^{\text{num pixels}} p_\theta(\bm{I}(i) \mid \bm{z}, \bm{c}(i))$\,.
        
    \vspace*{-1em}\paragraph{Monte Carlo approximation}
    Since \gls{OURS}'s decoder allows reconstructing single pixels, it is possible to compute a unbiased estimate of the image likelihood as 
    \begin{equation*}
        \hat{p}_\theta(\bm{I} \mid \bm{z}, \bm{c}) \approx  \frac{\text{num pixels}}{|S|} \sum_{s \in S} p_\theta(\bm{I}(s) \mid \bm{z}, \bm{c}(s))\,,
    \end{equation*}
    with $S$ a set of pixels selected uniformly at random from the image $\bm{I}$.
    This is useful for \gls{NERF}-based models due to the expensive rendering process (the scene function is evaluated many times for every pixel).
    Moreover, one can trade-off a bigger batch size (of scenes) for a smaller number of pixels used in the estimate---a useful property, since in our early experiments a bigger batch size (with fewer pixels) resulted in quicker (wall-clock time) model convergence and increased final performance.

    \vspace*{-1em}\paragraph{Using \Acrshort{GECO} to train pixel likelihood models}
    An alternative to setting $\beta$ manually is to use \glsreset{GECO}\gls{GECO} \cite{rezende2018geco}.
    \Gls{GECO} has shown excellent results in practice \cite{engelcke2019genesis,rezende2018geco}.
    It works by putting an explicit constraint on the desired per-pixel-likelihood, which unifies training of pixel and image generative models.
    We use \gls{GECO} to train our variant of \gls{GQN} and the \gls{GQN_baseline}.
    However, applying \gls{GECO} to \gls{NERF}-style models has its own issues: \gls{NERF} has two likelihood terms that are approximated with functions that are evaluated differently (\cref{app:subsec:nerf_details}), requiring different manually-chosen constraint values.
    Instead, we opted for $\beta$-annealing with \gls{ELBO}, which we use in all our experiments for \gls{OURS}.

\section{Experimental Details and Additional Results}
\label{app:exp_details}

    \begin{figure*}[htb]
        \includegraphics[width=\linewidth]{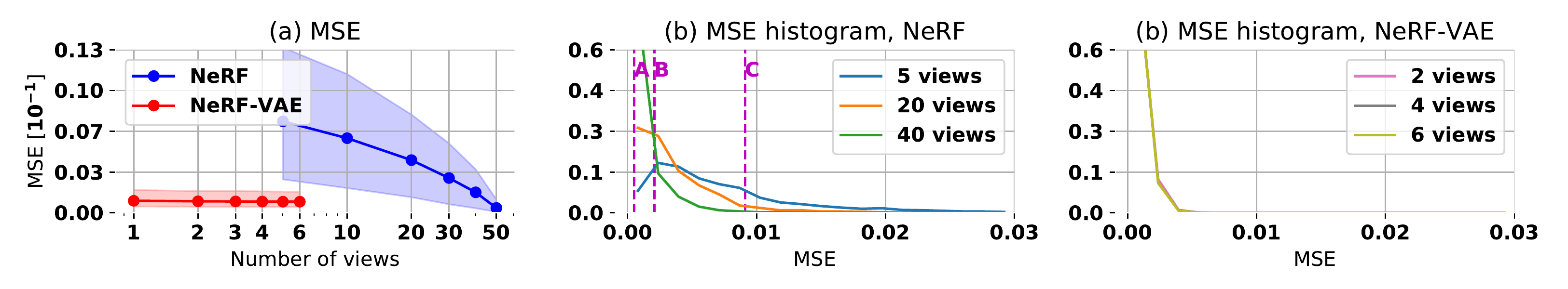}
         \includegraphics[width=\linewidth]{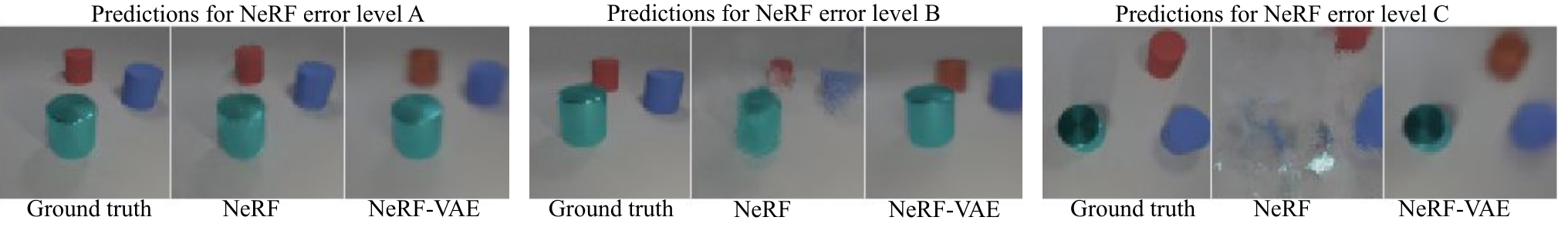}
         \caption{
        Error analysis of \gls{NERF} and a minimalistic version (\gls{MLP} decoder, no iterative inference) of \gls{OURS} on CLEVR data. \textbf{(a)}: \gls{MSE} decreases with increasing number of training (\gls{NERF}) and context (\gls{OURS}) views.
        We show the mean and 95\% percentiles across 100 test views averaged over 10 scenes and 10 seeds.
        \textbf{(b, c)}:
        \Gls{MSE} Histograms.
        Compared to \gls{OURS}, \gls{NERF} needs a large number of training views to consistently achieve low errors, and even then incurs a small number of larger errors.
        \textbf{Bottom}:
        An example scene, where views correspond to three error levels, indicated in (b), of \gls{NERF} trained on 40 views.
    }
    \label{fig:exp1_errors_and_densities_clevr}
    \end{figure*}  

    \subsection{Details: Comparison with NeRF}
    \label{app:exp1_details}
    
    For training \gls{OURS}, we evaluate the \gls{ELBO} of \cref{eq:elbo} by subsampling $512$ rays and colours from the context images uniformly at random.
    We use a $128$ dimensional latent variable, Adam with a learning rate of $5^{-4}$ for $1^6$ iterations, and $\beta=11 {-6}$, which is annealed to $1^{-4}$ from iteration 40k to 140k.
    The decoder is the \gls{MLP} scene function, with 4 layers with 64 units each for the both density and colour part, and a skip connection to layer two.
    Rays are evaluated within the interval $[0,7]$, with 32 and 64 points for coarse and fine scene function respectively.
    As in \citet{mildenhall2020nerf}, the densities are perturbed with Gaussian noise, but with smaller standard deviation of $0.01$.
    In this version of \gls{OURS}, we do not apply a CNN to the latent variable.
    The encoder ResNet uses  $[(5, 2), (5, 1)]$ block groups.
    
    \gls{NERF} is trained with a batch size of $256$ rays, using Adam with learning rate $1^{-3}$, for $5^6$ iterations.
    We use the same scene function architecture as for \gls{OURS}, but evaluate the rays on more points, 128 and 256 for coarse and fine respectively.

    \subsubsection*{CLEVR data}
    In \cref{subsec:nerf_comp} we show the benefits of amortizing NERF in the Jaytracer dataset in terms of view prediction errors as a function of the number of input views.
    We repeat the experiment on CLEVR data, using the same settings, apart from extending the ray intervals to $[0.05,14]$.
    
    We carry out the same experiment on our CLEVR dataset and test on a dataset of 10 scenes with 100 views per scene. We observe the same pattern again in \cref{fig:exp1_errors_and_densities_clevr} (a). In fact, in this case \gls{OURS} is able to perform well with just observing one view, and no further benefits are seen from having more views. This is because in CLEVR all objects are seen with one single image which allows the inference model to encode the whole scene. On the other hand, \gls{NERF} still suffers when it uses few views and it needs up to 50 views to gets comparable MSE with \gls{OURS}. For example, when 20 views are used, there is still a heavy tail of prediction errors across the 100 views as seen in \cref{fig:exp1_errors_and_densities_clevr} (b). Upon closer inspection of the predictions, these large errors are often produced by views in which the camera has an extreme viewpoint (e.g.\ very high elevation, as shown in the error level C of  \cref{fig:exp1_errors_and_densities_clevr}), as these views are less likely to be within the range of input views.

\subsection{Details: Comparison with a Convolution-Based Generative Model}
\label{app:exp2_details}

    We use the Adam optimizer \cite{kingma2014adam} with learning rate $=3\times 10 ^{-4}$ for Jaytracer and GQN datasets and $=5\times10^{-4}$ for the CLEVR dataset.
    The batch size is 192 for \gls{OURS} and 256 for \gls{GQN_baseline}.
    The non-iterative \gls{OURS}s and \gls{GQN_baseline} are trained for $10^6$ iterations and the iterative models are trained for $5 \times 10^5$ iterations (with 7 iterative inference steps).
    The $\beta$ annealing starts at the beginning of training for GQN and after 20k iterations for CLEVR at a low value of $\beta$ (0 for GQN, final $\beta / 1024$ for CLEVR) and linearly increases across 100k iterations to the final value.
    The final $\beta$ values were $[0.064, 0.256, 1.024, 4.096, 16.384]$ for GQN; for CLEVR these values were multiplied by a factor of 10.
    The iterative versions of \gls{OURS} estimate the image likelihood with a random sample of 256 pixels taken across the four target views; the non-iterative versions use 512 random pixels.
    \Gls{GQN_baseline} is trained with \gls{GECO} instead of $\beta$ annealing to a likelihood threshold of $3.3, 3.95$ and $4.0$ on Jaytracer, GQN and CLEVR datasets, respectively.
    
    All models use the encoder described in \cref{app:arch_details} with the following hyperparameters.
    All \gls{OURS}s use [(3, 2), (7, 1)] block groups in the encoder; the \gls{GQN_baseline} uses [(5, 2), (5, 1)] block groups.
    The latent $\bm{z}$ is a $128$-dimensional vector when the \gls{MLP} scene function is used.
    For the attentive scene function we use $8 \times 8 \times 128$ feature map.
    The \gls{MLP} used to predict the parameters of the posterior in \cref{eq:encoder} has 2 layers of 256 units each.
    We also apply an \gls{MLP} or a \gls{CNN} on the latent $\bm{z}$ before passing it into the scene function; the \gls{MLP} has one layer of 256 on Jaytracer and GQN data and two layers of 128 units each on CLEVR.
    The attentive function use a \gls{CNN} of three layers of 64 units each with kernel size 3 on all datasets.
    
    The \gls{MLP} scene functions use 2 density layers and 4 colour layers with 256 units each.
    The attentive scene function use 3 density layers and 3 colours layers with 256 units each.
    The final density and colour are predicted with a 2-layer \gls{MLP} from the intermediate features.

\subsection{Details: Interpolations \& Generalization on GQN data}
\label{app:gqn_cheating}

\begin{center}
    \includegraphics[width=\linewidth]{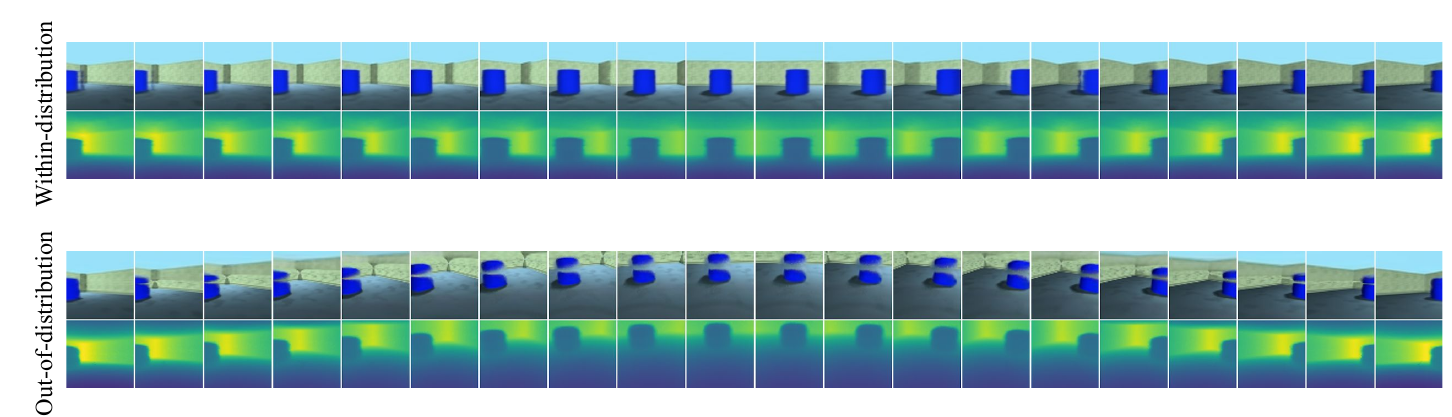}
    \begin{minipage}{0.3\linewidth}
        \captionof{figure}{
            \textbf{Top}:
            View interpolations on GQN data with \gls{OURS} where the scene function has access to orientations.
            Within-distribution interpolations look as expected, but out-of-distribution evaluations suffer from colour artifacts (depth estimates are unaffected). Compare with \cref{fig:exp2_out_of_plane_cameras} where the model has no access to orientations.
        }
        \label{fig:exp2_nerf_vae_ood_cameras_with_orientations}
    \end{minipage}
    \hfill
    \begin{minipage}{0.3\linewidth}
        \includegraphics[width=\linewidth]{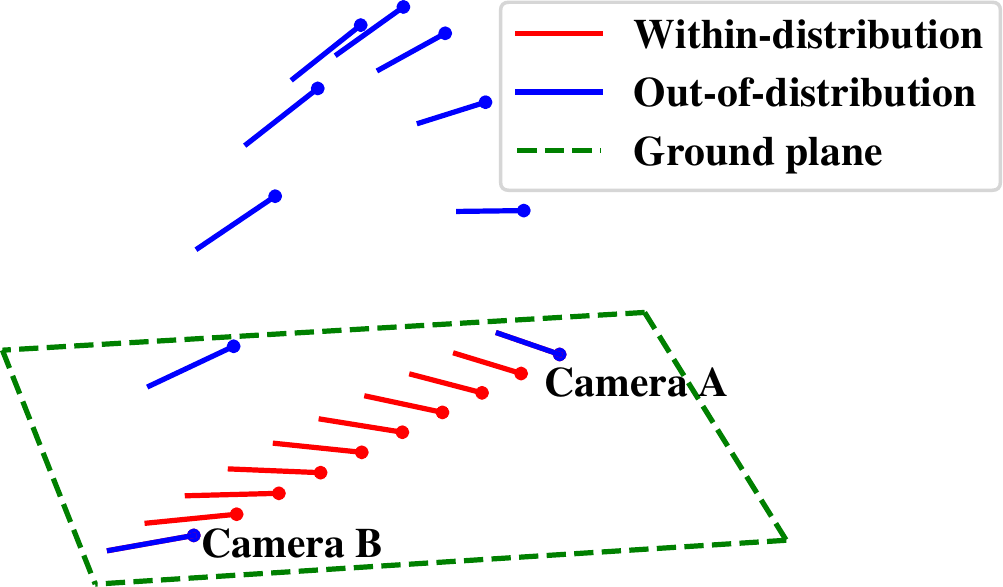}
    \end{minipage}
    \begin{minipage}{0.3\linewidth}
        \captionof{figure}{
            \textbf{Left}:
            Illustration for within and out-of-distribution camera trajectories for two representative cameras in the GQN dataset, used in \cref{fig:exp2_out_of_plane_cameras}.
            While within-distribution cameras are linear interpolations between the two endpoint positions and orientations within the ground plane, out-of-distribution cameras are additionally lifted of the ground and face downwards slightly.
        }
        \label{fig:exp2_camera_trajectories}
    \end{minipage}
\end{center} 
    \cref{fig:exp2_camera_trajectories} shows an illustration of how out-of-distribution camera trajectories are generated.

    We replicate \gls{OURS}'s out-of-distribution interpolations from \cref{fig:exp2_out_of_plane_cameras} with orientation inputs
    in \cref{fig:exp2_nerf_vae_ood_cameras_with_orientations} 
    While within-distribution interpolations look as expected, the colours  in the out-of-distribution interpolations are distorted---presumably as certain points in the scene function are evaluated at unseen orientations.
    Note that the densites, as depicted in the depth maps, are not affected by this problem, as densities are computed from positions only.

\subsection{Generalization on CLEVR}
\label{app:clevr_generalization}

We now investigate generalization on CLEVR data.
We focus on two aspects: 1) Comparing \gls{OURS} to \gls{GQN_baseline} on out-of-distribution camera views, 2) comparing the attention-based scene function compared to the \gls{MLP} on out-of-distribution number of objects.

\vspace*{-1em}\paragraph{Novel camera viewpoints} As with the GQN data, we expect \gls{OURS} to also generalize better to unseen viewpoints in CLEVR compared to the convolutional baseline \gls{GQN_baseline}.
To evaluate this we create a dataset with 20 scenes and 64 views.
Four views are sampled as per the camera distribution of original training dataset, and used as the context views.
The remaining 60 views are out-of-distribution, where 20 views interpolate between different elevations (at distance 4 to origin), 20 views where we interpolate between different azimuths (at distance 4 to origin) and 20 views where we interpolate distance to the origin between 2 and 6 (compared to sampled from $[6,8]$ during training).
We train the (\textsc{ii+attn})-variant of \gls{OURS}  and compare it with \gls{GQN_baseline}. 
\cref{fig:clevr_generalization_examples} (left), and \cref{fig:clevr_generalization_results} (a) show that \gls{OURS} achieves higher quality image outputs, as also indicated by lower \gls{MSE} and \gls{KL}.

\vspace*{-1em}\paragraph{Number of objects} 
We generate a test dataset (20 scenes, 60 views) that instantiates four or five objects instead of one to three in the training dataset.
We train and compare \gls{OURS} with both the \gls{MLP} and the attention-based scene function.
\cref{fig:clevr_generalization_examples} (right), and \cref{fig:clevr_generalization_results} (b) show that the attention-based scene function is able to generalize to larger number of objects, whereas the \gls{MLP} scene function misses objects.
This also results in lower \gls{MSE}.
We hypothesize this attractive feature of the attentive scene function is due to its spatial latent representation---enabling the ability to handle objects in scenes irrespective of the spatial arrangement and quantity seen during training.
    \begin{center}
        \includegraphics[width=.9\linewidth]{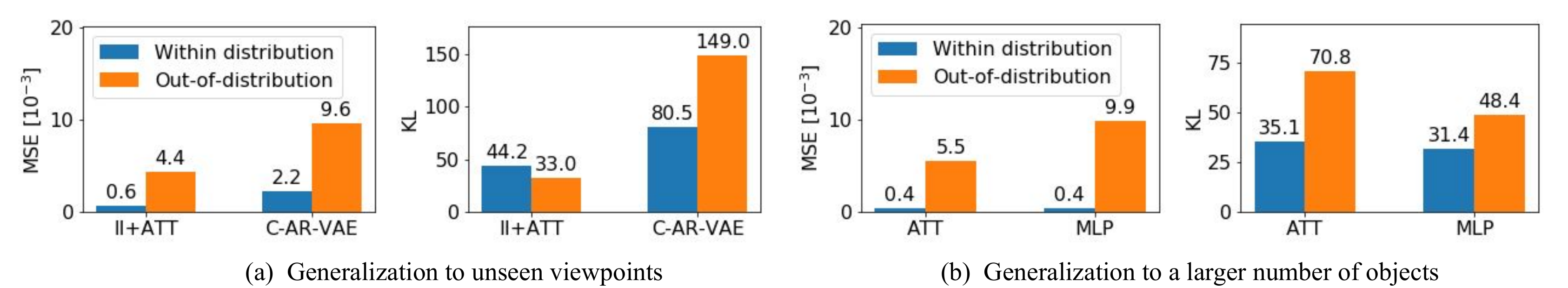}\hspace{0.1cm}
        \captionof{figure}{
            \textbf{(a)} Comparison of \gls{OURS} and \gls{GQN_baseline} on out-of-distribution camera viewpoints. \gls{OURS} obtains lower \gls{MSE} and \gls{KL}.
            \textbf{(b)} Comparison of \gls{MLP} and \textsc{attn} scene functions on out-of-distribution number of objects. \textsc{Attn} improves generalization.
            Median \gls{MSE} and \gls{KL} over 5 seeds.
        }
        \label{fig:clevr_generalization_results}
    \end{center}
    
    \begin{center}
        \includegraphics[width=.90\linewidth]{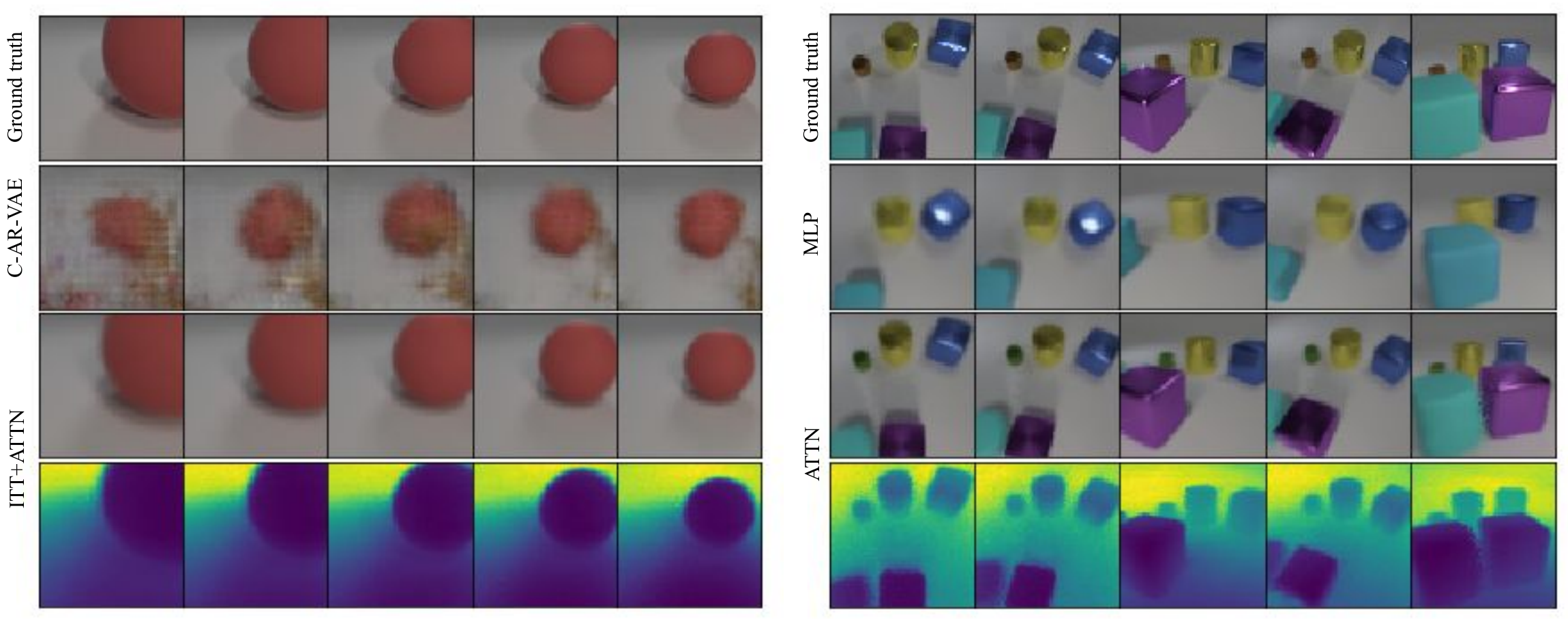}\hspace{0.1cm}
        \captionof{figure}{
        \textbf{Left:} Generalization to out-of-distribution camera views. While \gls{GQN_baseline} fails to produce plausible predictions, \gls{OURS} generalizes well.
        \textbf{Right:} Generalization to larger number of objects. The \gls{MLP} scene function misses objects, while \textsc{attn} captures all objects.
        }
        \label{fig:clevr_generalization_examples}
    \end{center}

\end{document}